\documentclass[12pt]{article}

\usepackage{newtxtext,newtxmath}

\usepackage{graphicx}

\usepackage[letterpaper,margin=1in]{geometry}

\renewenvironment{abstract}
	{\quotation}
	{\endquotation}

\date{}

\makeatletter
\renewcommand{\fnum@figure}{\textbf{Figure \thefigure}}
\renewcommand{\fnum@table}{\textbf{Table \thetable}}
\makeatother

\usepackage{scicite}

\usepackage{url}

\usepackage[all]{xy} 
\newtheorem{theorem}{Theorem}[section]
\newtheorem{proposition}[theorem]{Proposition}
\newtheorem{lemma}[theorem]{Lemma}
\newtheorem{corollary}[theorem]{Corollary}
\newtheorem{definition}[theorem]{Definition}

\def\scititle{
	Quotient Complex Transformer (QCformer) for Perovskite Data Analysis
}
\title{\bfseries \boldmath \scititle}

\author{
	Xinyu You$^{1}$,
	Xiang Liu$^{1}$,
	Chuan-Shen Hu$^{1}$,
        Kelin Xia$^{1\ast}$,
         Tze Chien Sum$^{2\ast}$\and
	\small$^{1}$Division of Mathematical Sciences, School of Physical and Mathematical Sciences,\and
    \small Nanyang Technological University, Singapore 637371, Singapore;\and
	\small$^{2}$Division of Physics \& Applied Physics,
School of Physical and Mathematical Sciences, \and
\small Nanyang
Technological University, Singapore 637371, Singapore.\and
	\small$^\ast$Corresponding author. Email: xiakelin@ntu.edu.sg; tzechien@ntu.edu.sg.\and
}

\begin{document} 

\maketitle

\begin{abstract} \bfseries \boldmath
The discovery of novel functional materials is crucial in addressing human’s greatest challenge of sustainable energy generation and climate change. 
Among the various types of materials, hybrid organic-inorganic perovskites (HOIPs) have attracted immense attention due to their exceptional optoelectronic properties for photvoltaics. 
 Recently, geometric deep learning, particularly graph neural networks (GNNs), has shown tremendous potential in predicting material properties and guiding material design. However, traditional GNNs often struggle to fully capture the information on the periodic structures and higher-order interactions that are prevalent in such material systems. To address these limitations, we propose a novel material structure representation based on quotient complexes (QCs) and introduce the Quotient Complex Transformer (QCformer) for material property prediction. In this approach, a material structure is modeled as a quotient complex, which assimilates both pairwise and many-body interactions through simplices of varying dimensions while incorporating material periodicity via a quotient operation. Our model utilizes efficient higher-order features defined on simplices and processes them using a simplex-based Transformer module. We pretrain the QCformer on widely used material benchmark datasets, including the Materials Project and JARVIS, and evaluate its performance in predicting the bandgap property of two HOIP datasets. Our results show that QCformer surpasses state-of-the-art models, underscoring its effectiveness for perovskite materials. Notably, the quotient complex material structure representation and its corresponding QCformer predictive model adds an invaluable tool to the existing toolkit for predicting properties of perovskite materials.
\end{abstract}


\section{Introduction}
\label{intro}
Perovskite materials have emerged as highly promising candidates for photovoltaic absorbers in solar cells in recent decades. Compared to traditional silicon-based solar cells, perovskite-based solar cells have lower fabrication cost and higher conversion efficiencies with the tandem configuration. As reported by National Renewable Energy Laboratory in 2025, the highest efficiency perovskite-silicon tandem solar cell and all-perovskite tandem solar cells have reached power conversion efficiency of 34.6\% and 30.1\%, respectively, compared to 22\% average for silicon-based solar panels. These advancements highlight the immense potential of perovskite solar cells for disrupting photovoltaic technology. However, achieving further improvements relies heavily on the discovery and design of new materials. Even materials composed of simple elements can occupy a huge chemical space, making traditional trial-and-error-based materials discovery both time-consuming and costly \cite{pilania2013accelerating}. 

Traditional computational models, like density functional theory and molecular dynamics, are computationally expensive. The growing availability of materials data has spurred the rise of AI-based material informatics, offering efficient and powerful tools for property prediction and design. 
Initiated in 2011, the Materials Genome Initiative (MGI) marked a milestone in AI-driven material data analysis \cite{national2011materials}, contributing significantly to the establishment of major materials databases, including the Materials Project \cite{jain2013materials}, JARVIS \cite{choudhary2020joint}, NOMAD, Aflowlib, and OQMD, as well as to the development of material AI models.

In general, AI-driven material models can be classified into two categories: feature-based machine learning models and end-to-end deep learning models. 
Material descriptors or fingerprints play a key role in feature-based material machine learning models. These descriptors are typically obtained from the physical and structural properties of materials, including ionization potential, ionic polarizability, electron affinity, Pauling electronegativity, valence orbital radii, HOMO and LUMO, atomic/ionic/orbital radii, tolerance and octahedral factors, packing factor, crystal structure measurements, surface/volume features, among others. A material fingerprint is a long vector composed of systematically generated features primarily derived from material structures. Prominent material fingerprints include the Coulomb matrix \cite{himanen2020dscribe}, Ewald sum matrices \cite{rupp2012fast}, many-body tensor representation (MBTR) \cite{huo2017unified}, smooth overlap of atomic positions (SOAP) \cite{bartok2013representing}, and atom-centered symmetry functions (ACSF) \cite{behler2011atom}.

Geometric deep learning models \cite{atz2021geometric,bronstein2017geometric,bronstein2021geometric,masci2015geodesic}, designed to analyze non-Euclidean data such as graphs, networks, and manifolds, have been widely applied in material data analysis.
In particular, various GNN models have been developed for material property analysis, which can be broadly classified into three categories: potential-energy-based GNNs, chemical-composition-based GNNs and geometric GNNs. 
The potential-energy-based GNNs, which include SchNet~\cite{schmidt2017predicting}, MEGNet~\cite{chen2019graph}, ALIGNN \cite{choudhary2021atomistic} and NequIP~\cite{batzner20223}, utilize specially designed GNN architectures to approximate potential energy, which is then used to predict material properties.
The chemical-composition-based GNNs, including CGCNN \cite{xie2018crystal}, iCGCNN \cite{park2020developing}, GATGNN \cite{louis2020graph}, CYATT \cite{schmidt2021crystal}, CrabNet \cite{wang2021compositionally}, MatDeepLearn \cite{fung2021benchmarking}, Matformer \cite{yan2022periodic}, ComFormer\cite{yan2024complete} and SIGNNA \cite{na2023substructure}, construct deep learning architectures directly on chemical formulas or chemical structures. Additionally, some approaches convert crystal properties into textual representations and use text-based models, such as CHGNet \cite{deng2023chgnet} and fine-tuned LLaMA-2 \cite{gruver2024fine}.
The geometric GNNs, including NequIP \cite{batzner20223}, Equivariant Networks for Crystal Structures \cite{kaba2022equivariant}, M3GNet \cite{chen2022universal}, FAENet \cite{duval2023faenet} and CHGNet \cite{deng2023chgnet} , which represent materials as geometric graphs with atoms embedded in 3D Euclidean space. They focus on invariant and equivariant transformations of geometric attributes under physical and hierarchical symmetries \cite{kaba2022equivariant,duval2023hitchhiker}. 
Among these, the ComFormer model achieves state-of-the-art performance in crystal property prediction by constructing E(3) periodic invariant crystal multigraph representations and incorporating periodic patterns in the node features. 

Despite these advancements, two major challenges remain in material geometric deep learning. First, effectively representing the periodic patterns in crystal structures is particularly challenging due to the inherent complexity of the periodic trends in materials \cite{batzner20223,fuchs2020se,keriven2019universal}. Historically, crystal structures are modeled as infinite graphs known as uniform nets \cite{wells1954geometrical,wells1977three}. Numerous studies have focused on identification of the nets that describe the underlying topology of crystal structures \cite{iwamoto1991inclusion,iwamoto1997mineralomimetic,furukawa2008control,o2012deconstructing,friedrichs2003three,delgado2003identification,o2008reticular}. These nets are not merely infinite graphs; they also possess translational actions that render them finite when factored out. This resulting finite graph is referred to as the quotient graph \cite{chung1984nomenclature,bader19973,eon1998geometrical,klee2004crystallographic}. 
Various theoretical investigations have explored quotient graphs, including the study of geometric relationships between quotient graphs and their underlying nets \cite{eon1998geometrical}, the application of covering space and homology theory from algebraic topology to analyze crystal quotient graphs \cite{sunada2012topological,sunada2012lecture}, the use of quotient graph-assisted algebraic methods to investigate crystal underlying nets and embeddings \cite{eon2016topological}, and the employment of the extended vector method to examine the reduction of quotient graphs \cite{eon2012symmetry}.
The quotient operation plays a pivotal role in capturing periodic patterns within the graph representation of materials. It identifies periodically equivalent atoms and edges in the infinite crystal graph representation, thereby producing the quotient graph \cite{chung1984nomenclature, bader19973}. The quotient graph model has been applied in recent GNN models for material data analysis\cite{xie2018crystal,chen2019graph,park2020developing,louis2020graph,schmidt2021crystal,choudhary2021atomistic,yan2022periodic}.
These GNN models typically use the $k$-nearest neighbors method to generate a large graph, which is then transformed into a quotient graph by gluing all repetitive atoms in different unit cells together. The resulting quotient graph is essentially a multi-graph, where the vertices represent atoms from the unit cell and the multi-edges represent atomic interactions both within and between unit cells. 

Another key issue in material geometric deep learning is the characterization of multiscale higher-order interactions. While graph-based topological representations are effective for capturing pairwise interactions, they often struggle to accurately describe higher-order or many-body interactions, which are prevalent in material data. Mathematically, a graph can be generalized to a simplicial complex, where higher-order interactions are geometrically represented as features on triangles, tetrahedrons, and higher-dimensional components. This forms the foundation of topological deep learning. Recently, simplicial-complex-based topological data analysis (TDA) methods have been proposed to handle data with higher-order interaction structures \cite{papillon2023architectures,hajijtopological}. Various topological deep learning (TDL) models have been developed, such as Simplicial Neural Networks (SNNs) \cite{ebli2020simplicial}, Hypergraph Attention Networks (HANs) \cite{kim2020hypergraph}, Cell Complexes Neural Networks (CXNs) \cite{hajij2020cell}, CW Networks (CWNs) \cite{bodnar2021weisfeiler}, Block Simplicial complex Neural Networks (BScNets) \cite{chen2022bscnets}, Simplicial Attention Networks (SAT) \cite{goh2022simplicial}, Simplicial Graph Attention Network (SGAT) \cite{lee2022sgat}, Higher-Order Attention Networks (HOANs) \cite{hajij2022higher} and others \cite{papillon2023architectures}. Compared to GNNs, TDL models make use of more general topological representations, such as simplicial complexes, hypergraphs, cell complexes, and combinatorial complexes, and define higher-order features on these structures. Additionally, the message-passing or aggregation modules in TDL are more complicated with neighbor strategies that include upper adjacency, lower adjacency, face relationships, coface relationships, and others. To the best of our knowledge, no TDL models have yet been applied to material data analysis.

Here, we propose a quotient-complex-based material representation and a novel model, the Quotient Complex Transformer (QCformer), for perovskite data analysis. In the QCformer model, a crystal structure is represented as a quotient complex, where pairwise interactions are embedded in the edges, and higher-order interactions are encoded in triangles and other higher-dimensional components. The QCformer model consists of multiple Simplex Transformer (Sformer) blocks, which use attention mechanism to facilitate message passing and update the embedding features within the crystal quotient complex. 

More importantly, we conducted experiments on two chemical systems: inorganic materials and hybrid organic-inorganic perovskite (HOIP) materials. Our model achieved state-of-the-art performance in predicting various properties of the Materials Project and JARVIS dataset, both of which consist of inorganic materials. To utilize the knowledge learned from these large-scale inorganic datasets, we pre-trained our model on JARVIS and transferred the learned embeddings to HOIP materials, enabling more accurate predictions despite the limited availability of HOIP data. In our evaluations, QCformer outperformed both traditional machine-learning models and GNN models, achieving superior accuracy in bandgap prediction and demonstrating exceptional capability in identifying novel 2D perovskite materials. Furthermore, we found that encoding three-body interactions using triangles significantly improved predictive performance across all tasks, highlighting the effectiveness of our quotient complex structure in capturing higher-order material relationships. These results reinforce the potential of QCformer to advance HOIP research, providing a powerful tool for accelerating the discovery and design of next-generation perovskite materials.

\section{Results}
\subsection{Quotient Complex Transformer (QCformer) }
\subsubsection{Quotient Complex Representation for Materials}
A crystal structure can be represented by a graph, where vertices are atoms and edges are atomic bonds. By gluing equivalent atoms together in the graph, that is, treating them as a single atom, we can construct a quotient graph. The quotient graph has been successfully used to represent crystal structures \cite{xie2018crystal,yan2022periodic,choudhary2021atomistic}. As a generalization of graphs, simplicial complexes encompass not only vertices and edges but also triangles, tetrahedrons and other higher-dimensional counterparts. These higher dimensional components naturally characterize higher-order or many-body interactions. Once suitable equivalence relation(s) are defined, a quotient complex (QC) can be constructed from a simplicial complex, by gluing the equivalent vertices/edges/higher-order simplices together.

\begin{figure}[ht]
	\centering
	\includegraphics[width=0.95\textwidth]{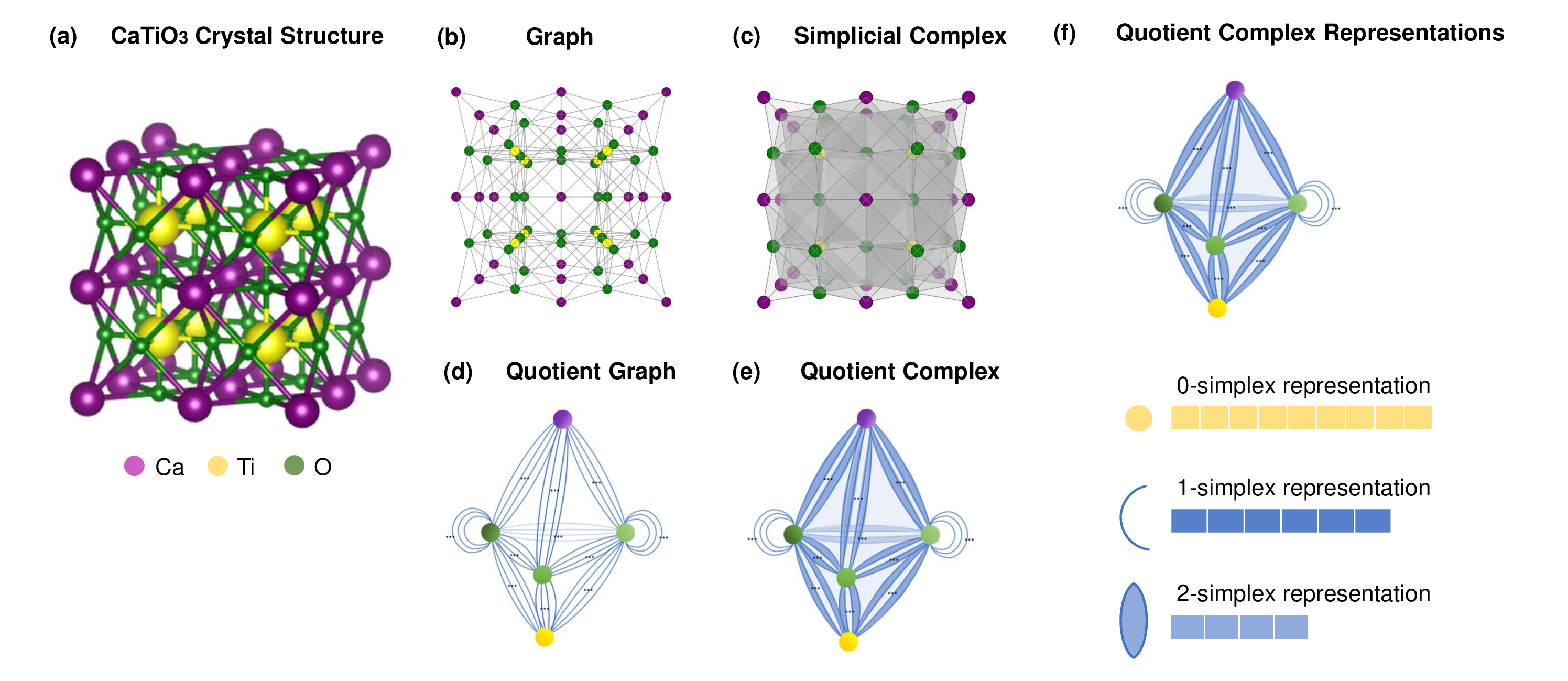}
	\caption{\textbf{Illustration of a finite crystal structure extended by relative topological representations.} Starting from the CaTiO$_3$
  crystal \textbf{(a)}, we transform it into a graph \textbf{(b)}, then into a simplicial complex \textbf{(c)}, and finally into their respective quotient versions \textbf{(d)}\textbf{(e)} by merging nodes of the same color. The finite cell structure is visualized using the VESTA program~\cite{momma2011vesta}. \textbf{(f)} The resulting quotient complex decomposes into 0‐simplices (nodes), 1‐simplices (edges), and 2‐simplices (triangles), each with its associated features.} 
	\label{fig:quotient-complex}
\end{figure}

The construction of crystal quotient graph and crystal quotient complex is demonstrated in Figure \ref{fig:quotient-complex}(a).
The crystal structure of the material $\text{CaTiO}_3$ can be visualized as an infinite graph in which vertices correspond to atoms and edges represent interactions between them. A finite crystal graph or simplicial complex can be extracted from this infinite crystal graph by considering only the atoms within a defined supercell (composed of multiple unit cells) or the $k$-nearest neighbors of atoms within the unit cell. For illustration, we consider a supercell-based graph composed of $2\times 2\times 2$ unit cells. By gluing vertices with same colors together, a quotient graph is constructed. In contrast, a simplicial complex can be constructed based on the points in the supercell. Using the same equivalence relations as in the quotient graph, a quotient complex can be constructed. This quotient complex encompasses multiple edges to account for pairwise interactions between distinct vertices, self-loops representing self-interactions of individual vertices, and triangles that indicate higher-order interactions involving three distinct vertices. Thus, the quotient complex can be seen as a generalization of the quotient graph.

Similar to how standard graph‐based methods assign node and edge attributes, our approach generalizes feature assignment to all simplexes in the structure, as illustrated in Figure \ref{fig:quotient-complex}(b). Both the resulting quotient‐complex connectivity and the assigned simplex features become inputs to QCformer, enabling the model to learn from richer, multi‐dimensional representations of the crystal structure.

\subsubsection{Simplex Transformer and QCformer}
A key element of our QCformer model is the Simplex Transformer (Sformer) module, which updates simplex representations by aggregating information from their neighbors and cofaces. We employ the standard scaled dot-product attention mechanism to build the aggregation function. Thus, our message-passing scheme consists of three steps: attention coefficient computation, message computation, and simplex feature updating. Denoting the constructed crystal quotient complex as $C$, for any $n$-simplex $\sigma_n\in C$, assume that $\tau_n$ is a neighbor of $\sigma_n$. Two $n$-simplices are considered neighbors if they both serve as faces of a shared $(n+1)$-simplex, referred to as a coface. Unlike simplicial complex attention models, in our quotient complex representation, two $n$-simplices that are neighbors can share multiple common cofaces. For instance, two vertices may share several edges connecting them. We denote the $s$-th common coface of $\sigma_n$ and $\tau_n$ by $(\sigma_n,\tau_n)_s$. Let ${\bf h}_{\sigma_n}$ be the feature vector for the $n$-simplex $\sigma_n$. Its query ${\bf q}$, key ${\bf k}$ and value ${\bf v}$ can be expressed as,
\begin{equation}
    {\bf q}_{\sigma_n}={\bf Q}_n {\bf h}_{\sigma_n}
\end{equation}
\begin{equation}
    {\bf k}_{\tau_n}={\bf K}_n {\bf h}_{\tau_n}, ~~{\bf k}_{(\sigma_n,\tau_n)_s}={\bf K}_{n+1} {\bf h}_{(\sigma_n,\tau_n)_s},
    {\bf v}_{\tau_n}={\bf V}_n {\bf h}_{\tau_n}, ~~{\bf v}_{(\sigma_n,\tau_n)_s}={\bf V}_{n+1} {\bf h}_{(\sigma_n,\tau_n)_s}
\end{equation}

where ${\bf Q}_n$,  ${\bf K}_n$, and ${\bf V}_{n}$ are the query, key, and value matrices for $n$-simplices, respectively. Matrices ${\bf K}_{n+1}$ and ${\bf V}_{n+1}$ are the key and value matrices for $(n+1)$-simplices.

Next, we compute the message from $\tau_n$ to $\sigma_n$ through $(\sigma_n,\tau_n)_s$,
\begin{equation}
    {\bf \alpha}_{(\sigma_n,\tau_n)_s}=\frac{[{\bf q}_{\sigma_n},{\bf q}_{\sigma_n}]\circ {\rm MLP}[{\bf k}_{\tau_n},{\bf k}_{(\sigma_n,\tau_n)_s}]}{\sqrt{2d({\bf q}_{\sigma_n})}}
\end{equation}
\begin{equation}
    {\bf m}_{(\sigma_n,\tau_n)_s}={\rm Sig}({\rm BN}(\alpha_{(\sigma_n,\tau_n)_s}))\circ {\rm MLP}([{\bf v}_{\tau_n},{\bf v}_{(\sigma_n,\tau_n)_s}])
\end{equation}
where $d({\bf q}_{\sigma_n})$ is the dimension of ${\bf q}_{\sigma_n}$, ${\bf \alpha}_{(\sigma_n,\tau_n)_s}$ is the attention coefficient vector of $\tau_n$ to $\sigma_n$ through their $s$-th common coface $(\sigma_n,\tau_n)_s$. The symbol $\circ$ represents the Hadamard product, and $[x,y]$ represents the concatenation of $x$ and $y$. $\rm Sig$ denotes the Sigmoid activation function, $\rm BN$ represents the Batch Normalization operation, and $\rm MLP$ is the Multi-Layer Perceptron used to compute the updated message. Here, we follow the Matformer model and use Sigmoid and Batch Normalization instead of the Softmax operation.

\begin{figure}[ht]
	\centering
	\includegraphics[width=0.95\textwidth]{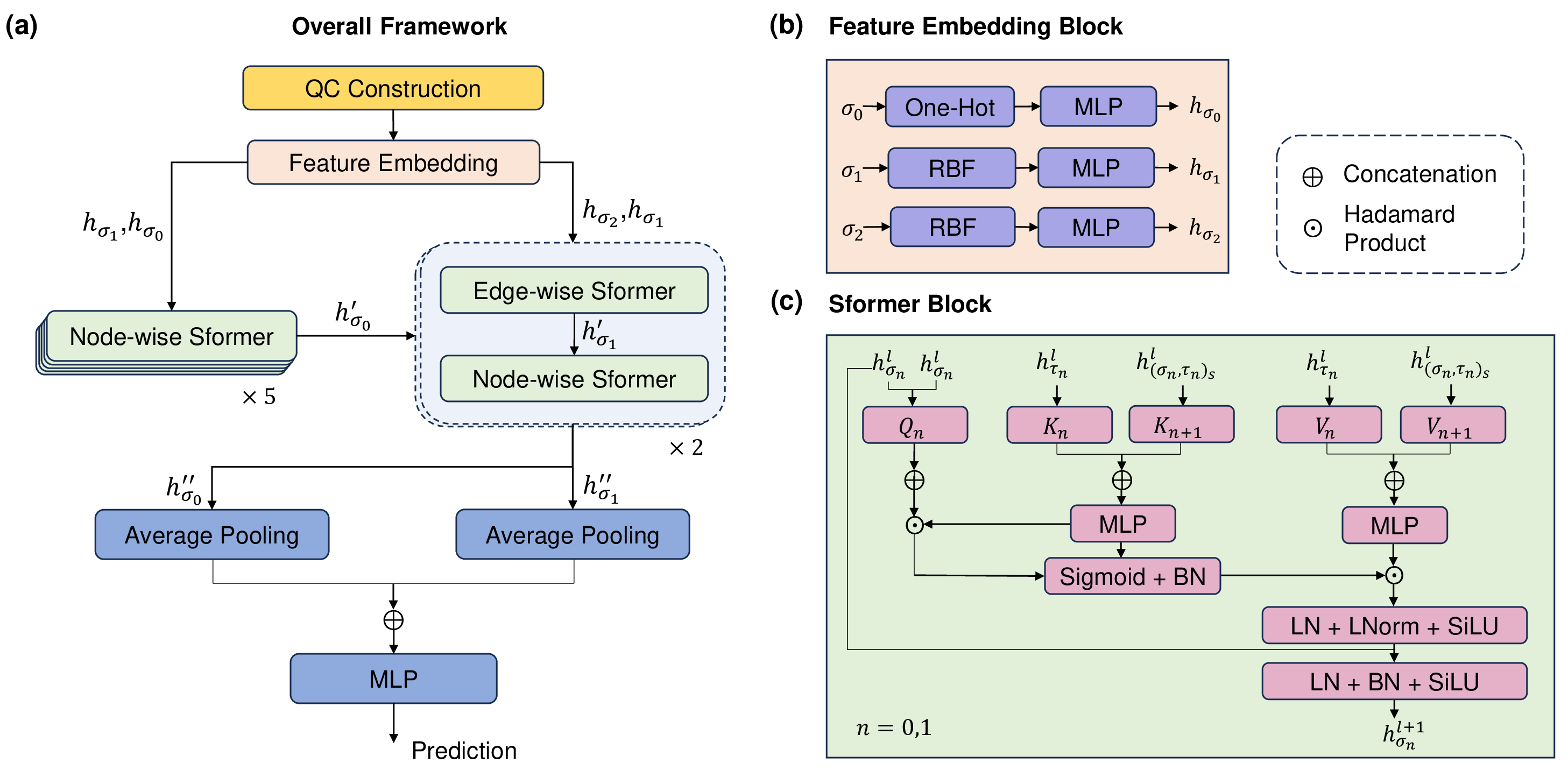}
	\caption{\textbf{Detailed illustration of the QCformer architecture.} {\bf (a)}: The overall framework of QCformer, where quotient complexes (QCs) are constructed and processed through feature embedding, message passing, and pooling modules. The final prediction is obtained via an MLP layer. {\bf (b)}:Feature embedding block in {\bf (a)}. The 0-simplex feature is embedded as a one-hot vector, following the approach in CGCNN \cite{xie2018crystal}. The features of 1-simplices and 2-simplices are embedded using radial basis functions (RBF). Finally, simplices of all dimensions are projected into the same hidden dimension through an MLP layer. {\bf (c)}: Detailed Simplex Transformer (Sformer) block in {\bf (a)}. Note that each $n$-simplex $\sigma_n$ is updated by aggregating information from its neighbors ($n$-simplices) $\tau_n$ and cofaces ($(n+1)$-simplices) $(\sigma_n,\tau_n)_s$ and higher dimensional simplices are updated to be inputs for the updating of lower dimensional simplices.}
	\label{fig:model}
\end{figure}

Finally, we compute the updated feature vector of $\sigma_n$ in $(l+1)$-th layer.
\begin{equation}
    {\bf m}^{(l)}_{{\bf \sigma}_n}=\sum_{\tau_n\in N({\sigma}_n)}\sum_{s}{\rm SiLU}({\rm LNorm}({\rm LN}({\bf m}^{(l)}_{(\sigma_n,\tau_n)_s})))
\end{equation}
\begin{equation}
    {\bf h}^{(l+1)}_{{\bf \sigma}_n}={\bf h}^{(l)}_{\sigma_n}+{\rm SiLU}({\rm BN}({\rm LN}({\bf m}^{(l)}_{\sigma_n})))
\end{equation}
where $N(\sigma_n)$ is the set of all neighbors of $\sigma_n$. $\rm SiLU$ is the Sigmoid Linear Unit activation function. $\rm BN$ is the Batch Normalization operation. $\rm LN$ is the Linear
transformation. An illustration of our Simplex Transformer (Sformer) module is demonstrated in Figure \ref{fig:model}(b).

We adopt general GNN modules from Matformer \cite{yan2022periodic} in our QCformer, utilizing a message-passing scheme. In standard GNN models, message flow typically occurs from the edge level to the node level. However, with higher-order components, it is natural to extend message passing from triangles to edges. The overall framework is illustrated in Figure 2(a). This generalized higher-order message-passing scheme can be applied to any simplicial complex of any order, provided that boundary relations are properly defined. The details of message passing are discussed in Section  \ref{methods:model}.



\subsection{QCformer for Material Property Prediction }
\subsubsection{Model Performance}
We test our model on two benchmark datasets: Materials Project \cite{jain2013commentary} and JARVIS \cite{choudhary2020joint}. 
Specifically, we use the 2018.6.1 version of the Materials Project dataset and the 2021.8.18 version of the JARVIS-DFT dataset. For the Materials Project dataset, we evaluate our model on four crystal property prediction tasks, including Formation Energy, Band Gap, Bulk Moduli and Shear Moduli. For the JARVIS dataset, we consider five crystal property prediction tasks, including Formation Energy, Total Energy, Bandgap(OPT), Bandgap(MBJ) and ${\text E}_{\text{hull}}$. We follow Matformer \cite{yan2022periodic} and use the same training, validation and test set for all the prediction tasks. 
The mean absolute error (MAE) is used to qualify the model performance. Detailed training, validation and test set can be found in Supplementary materials \ref{Appendix: Dataset Details}.

\paragraph{The Materials Project dataset} We first evaluate our QCformer model using the Materials Project 2018.6.1 dataset, which contains 69,239 crystals. The results, summarized in Table \ref{table:mp-result}, show that QCformer outperforms all other models on three out of four tasks and achieves performance comparable to the second-best model on the Formation Energy task. Specifically, for Bulk Moduli and Shear Moduli, QCformer reduces the MAE of the second-best model by 13.16\% and 7.81\%,  respectively. Moreover, QCformer achieves the best performance on Bulk Moduli and Shear Moduli with 4,664 training samples, and on Formation Energy and Band Gap with 60,000 training samples. This demonstrates the model's robustness across prediction tasks involving datasets of varying sizes. 
\begin{table}[ht]
	\caption{\textbf{Model performance (MAE) comparison between our model (QCformer) and existing models on the Materials Project dataset.} The best results are shown in {\bf bold}. The units are eV/atom, eV, log(GPa), log(GPa) for Formation Energy, Band Gap, Bulk Moduli and Shear Moduli, respectively.}
	\label{table:mp-result}
	\centering
		\begin{tabular}{l|cccc}
			\hline
			Method &Formation Energy & Band Gap & Bulk Moduli & Shear Moduli\\
			\hline
			CGCNN \cite{xie2018crystal}  &  0.031& 0.292& 0.047 & 0.077 \\
			SchNet \cite{schmidt2017predicting} &  0.033& 0.345& 0.066 & 0.099  \\
			MEGNET \cite{chen2019graph} &  0.030& 0.307& 0.060& 0.099  \\
			GATGNN \cite{louis2020graph} & 0.033 & 0.280& 0.045 & 0.075  \\
			ALIGNN \cite{choudhary2021atomistic} & 0.022 & 0.218& 0.051& 0.078  \\
			Matformer \cite{yan2022periodic} & 0.021& 0.211& 0.043& 0.073\\
            ComFormer \cite{yan2024complete} & 0.018& 0.193& 0.038& 0.064\\
			{\bf QCformer} &   {\bf 0.018} & {\bf 0.191} & {\bf 0.033}& {\bf 0.059}\\
			\hline			
		\end{tabular}
\end{table}

\paragraph{The JARVIS dataset} We evaluate our QCformer model on the JARVIS 2021.8.18 dataset, which comprises 55,722 crystals. The results, presented in Table \ref{table:jarvis-result}, demonstrate that QCformer outperforms all other models on the prediction tasks of Total Energy, Bandgap(OPT), and Bandgap(MBJ). Notably, QCformer reduces the MAE of the second-best model by 7.38\% for Bandgap(OPT) and 7.69\% for Bandgap(MBJ).

\begin{table}[ht]
	\caption{\textbf{Model performance (MAE) comparison between our model (QCformer) and existing models on the JARVIS dataset.} The best results are shown in {\bf bold}. The units are eV/atom, eV/atom, eV, eV, eV for Formation Energy, Total Energy, Bandgap(OPT), Bandgap(MBJ) and ${\text E}_{\text{hull}}$,  respectively.}
	\label{table:jarvis-result}
	\centering
	\begin{tabular}{l|ccccc}
		\hline
		Method &Formation Energy & Total Energy & Bandgap(OPT) & Bandgap(MBJ)& ${\text E}_{\text{hull}}$\\
		\hline
		CFID \cite{PhysRevMaterials.2.083801}   &  0.14 & 0.24 & 0.30 & 0.53 &0.22\\
		CGCNN \cite{xie2018crystal}  &  0.063& 0.078& 0.20 & 0.41 &0.17\\
		SchNet \cite{schmidt2017predicting} &  0.045& 0.047& 0.19 & 0.43  &0.14\\
		MEGNET \cite{chen2019graph} &  0.047& 0.058& 0.145& 0.34  &0.084\\
		GATGNN \cite{louis2020graph} & 0.047 & 0.056& 0.17 & 0.51  &0.12\\
		ALIGNN \cite{choudhary2021atomistic} & 0.0331& 0.037& 0.142& 0.31  &0.076\\
		Matformer \cite{yan2022periodic} & 0.0325& 0.035&0.137&0.30 & 0.064\\
        ComFormer \cite{yan2024complete} & {\bf 0.0272}& \textbf{0.029}& 0.122& 0.26& {\bf 0.044}\\
		{\bf QCformer} &   0.0290 & {\bf 0.029} & {\bf 0.113}& {\bf 0.24}&  0.053\\
		\hline			
	\end{tabular}
\end{table}




\subsection{QCformer for Perovskite Bandgap Prediction}
\subsubsection{Model Performance}
We pretrained QCformer model using 55722 data points from the JARVIS dataset on the Bandgap (OPT) task to generate useful graph embeddings. The pretrained model is then applied to downstream tasks for predicting hybrid organic-inorganic perovskites bandgap.
We evaluate the performance on two datasets, one is hybrid organic-inorganic perovskites (HOIP) proposed by Kim et al. \cite{kim2017data}, the other is 2D hybrid organic-inorganic perovskites (HOIP2D) proposed by Laboratory of New Materials for Solar Energetics (NMSE)\cite{marchenko2020database}. Dataset details can be found in Supplementary materials \ref{Appendix: Dataset Details}.


\paragraph{HOIP dataset}
For HOIP dataset, we use 10-fold cross-validation and provide coefficient of
determination (COD), pearson correlation coefficient (PCC), mean absolute error (MAE), mean squared error (MSE), and root mean squared error (RMSE) for comparison with other baseline models.

The results in Table \ref{table:HOIP} show that our model not only outperforms traditional machine‐learning methods, but also achieves better accuracy than contemporary GNN approaches. For comparison, TPD-GBT, PRC$_a$-GBT and PH$_a$-GBT \cite{anand2022topological} are all gradient‐boosted tree (GBT) models, each enhanced by different descriptor sets—traditional perovskite descriptors (TPD), persistent Ricci curvature (PRC), or persistent homology (PH), respectively. While incorporating topological features such as PRC and PH already improves predictive performance over purely descriptor‐based ML, GNN methods typically yield even better results. 

Our approach, however, utilizes topological representations within simplicial‐complex neural networks, providing further gains beyond both GBT‐based and GNN‐based techniques. In particular, we observe a 30.59\% improvement in MAE over MEGNet, accompanied by a COD of 0.9829 and a PCC of 0.9916. These findings underscore the effectiveness of merging topological insights with neural architectures for accurately modeling bandgap properties.

\begin{table}[ht]
\centering
\caption{\textbf{Model performance comparison between our model(QCformer) and existing models on the HOIP dataset.} The comparison includes machine learning based models (TPD-GBT, PRC$_a$-GBT and PH$_a$-GBT), and GNN models (SchNet and MEGNet). The best results are shown in \textbf{bold}. The units are eV, eV$^2$ and eV for MAE, MSE and RMSE.}
\label{table:HOIP}
\begin{tabular}{l|c|c|c|c|c}
\hline
Method & COD & PCC & MAE & MSE & RMSE \\
\hline
TPD-GBT \cite{anand2022topological}    & 0.8270 & ---    & 0.3770 & 0.2010 & 0.4483 \\
PRC$_a$-GBT \cite{anand2022topological} & 0.9068 & 0.9531 & 0.2620 & 0.1083 & 0.3288 \\
SchNet \cite{schmidt2017predicting}    & 0.9274 & 0.9656 & 0.2112 & 0.0792 & 0.2800 \\
PH$_a$-GBT \cite{anand2022topological} & 0.9372 & 0.9689 & 0.2032 & 0.0729 & 0.2697 \\
MEGNet \cite{chen2019graph}     & 0.9673 & 0.9840 & 0.1396 & 0.0370 & 0.1924 \\
\textbf{QCformer} & \textbf{0.9829} & \textbf{0.9916} & \textbf{0.0969} & \textbf{0.0197} & \textbf{0.1398} \\
\hline
\end{tabular}
\end{table}

\paragraph{HOIP2D dataset}
For HOIP2D dataset, we use a five‐fold cross‐validation protocol, reporting the coefficient of determination (COD), Pearson correlation coefficient (PCC), mean absolute error (MAE), and root mean squared error (RMSE) as valuation 
 metrics. To ensure a fair comparison, we train and test our model on a randomly selected subset of 624 data points (out of 716 compounds).

As summarized in Table \ref{table:NMSE}, our model outperforms a range of GNN‐based methods, including GCN, ECCN, CGCNN, TFGNN, SIGNNA and SIGNNA$_c$. Among these, SIGNNA$_c$ attains the best performance, primarily by separating organic and inorganic sub‐structures. By incorporating topological features, our method achieves a MAE of 0.0754, approximately 9.2\% lower than SIGNNA$_c$. However, in general, GNN methods appear limited by the relatively small training set, which negatively impacts their performance compared to GBT models.

Beyond GNNs, our approach also surpasses several descriptor‐driven machine‐learning methods. For instance, the smooth overlap of atomic positions (SOAP) descriptor‐based GBT model and the geometric data analysis (GDA)‐based GBT model (which utilizes element‐specific density fingerprints)\cite{hu2024geometric} both exhibit higher errors than ours. Specifically, our model attains an RMSE of 0.1038 and a PCC of 0.9590, indicating fewer outliers and a strong positive correlation between predictions and ground truths. It is noteworthy, however, that our MAE does not quite match that of the GDA‐GBT model. This is a limitation likely from the relatively large number of parameters in our model combined with a comparatively small dataset, which increases the risk of overfitting.

\begin{table}[ht]
\centering
\caption{\textbf{Model performance comparison between our model(QCformer) and existing models on the HOIP2D dataset.} The comparison includes machine learning based models (SOAP-KRR, SOAP-MLM1, and GDA(IF)-GBT), and GNN models (GCN, ECCN, CGCNN, TFGNN, SIGNNA and SIGNNA$_c$).  The second column shows the numbers of materials involved in different experiments. The best results are shown in \textbf{bold}. The units are eV and eV for MAE and RMSE.}
\label{table:NMSE}
\begin{tabular}{l|c|c|c|c|c}
\hline
Method & Number of Compounds & COD & PCC & MAE & RMSE \\
\hline
SOAP-KRR \cite{mayr2021global}  & 445 & 0.6700 & ---    & 0.1250 & 0.2530 \\
SOAP-MLM1 \cite{marchenko2020database}  & 515 & 0.9005 & ---    & 0.1030 & 0.1360 \\
GCN \cite{na2023substructure}       & 624 & 0.6030 & ---    & 0.1910 & ---    \\
ECCN  \cite{na2023substructure}     & 624 & 0.3170 & ---    & 0.2240 & ---    \\
CGCNN \cite{na2023substructure}     & 624 & 0.6450 & ---    & 0.1750 & ---    \\
TFGNN \cite{na2023substructure}     & 624 & 0.6650 & ---    & 0.1620 & ---    \\
SIGNNA \cite{na2023substructure}    & 624 & 0.9080 & ---    & 0.0920 & ---    \\
SIGNNA$_c$ \cite{na2023substructure} & 624 & \textbf{0.9270} & ---    & 0.0830 & ---    \\
GDA(HF)-GBT  \cite{hu2024geometric}     & 624 & 0.8645 & 0.9316 & 0.0936 & 0.1400 \\
GDA-GBT  \cite{hu2024geometric}         & 624 & 0.9101 & 0.9548 & 0.0695 & 0.1136 \\
GDA(IF)-GBT  \cite{hu2024geometric}     & 624 & 0.9157 & 0.9582 & \textbf{0.0681} & 0.1114 \\
QCformer & 624 & 0.9179 & \textbf{0.9590} & 0.0754 & \textbf{0.1038} \\
\hline
\end{tabular}
\end{table}

\subsubsection{Bandgap Prediction For New Perovskite Materials}
Perovskite‐based solar cells have recently emerged as a promising direction for more eco‐friendly and sustainable energy generation. This motivates extensive research into the properties of novel perovskite materials. In order to assess the predictive power of our model and its potential application for perovskite material discovery, we applied it to a subset of 2D perovskites from the HOIP2D dataset that lacked known bandgap labels. 

Specifically, we trained our model on 624 data points from HOIP2D with known bandgaps, then tested on five unlabeled compounds indexed 110, 309, 315, 363, and 452. To provide a benchmark for our predictions, we performed GGA‐PBE‐based DFT calculations on these same compounds. Across the five compounds, our model achives a MAE of 0.3849 eV. as shown in Table \ref{table:new-bandgaps}. Although this MAE is higher than the result in Table \ref{table:NMSE}, this is expected, as a smaller dataset can lead to larger average prediction errors. A detailed comparison between the predicted bandgaps and the DFT results is provided in Section \ref{methods:new-prediction}.

\begin{table}[ht]
\caption{\textbf{Comparison of DFT and predicted bandgaps for five new compounds.} The first column and second column show the indices and chemical formulas. The third column presents calculated bandgaps by GGA‐PBE‐based DFT method. The fourth column showcase the predicted bandgaps by QCformer. }
\label{table:new-bandgaps}
\centering
\begin{tabular}{cccc}
\hline
No. & Formula & DFT bg (eV) & QCformer bg (eV) \\ \hline
110 & [(C$_6$H$_{11}$)PH$_3$]$_2$SnI$_4$ & 1.3540 & 2.2435 \\ 
309 & [C$_2$H$_5$NH$_3$]$_2$FeCl$_4$ & 2.7002 & 2.5898 \\ 
315 & Cs$_3$Sb$_2$I$_9$ & 1.5450 & 1.7782 \\ 
363 & [CH$_3$NH$_3$]$_2$FeCl$_4$ & 3.1661 & 3.1757 \\ 
452 & Tl$_3$Bi$_3$I$_9$ & 2.3780 & 1.6962 \\ \hline
\end{tabular}
\end{table}

\section{Discussion}
In this work, we propose two key contributions: a novel quotient complex-based material structure representation and a corresponding predictive model called the Quotient Complex Transformer (QCformer) for material property prediction. The quotient complex representation captures material structures by encoding pairwise interactions as edges and three-body interactions as triangles. Building on this representation, QCformer employs an attention-based message-passing mechanism, where each $n$-simplex is updated by aggregating information from its neighboring $n$-simplices and cofaces ($(n+1)$-simplices). Higher-dimensional simplices contribute to updating lower-dimensional simplices, enabling hierarchical information flow. 
We validate our approach on two benchmark datasets—the Materials Project and JARVIS datasets—achieving state-of-the-art results on seven material property prediction tasks. 
Furthermore, we evaluate our model on the HOIP datasets to assess its applicability beyond traditional material benchmarks. Our results demonstrate that QCformer effectively captures the underlying structural patterns that govern their electronic properties, which highlights the potential of our approach in accelerating the discovery of novel perovskite materials with desirable bandgaps.

\section{Materials and Methods}

\subsection{Crystal Structure}
A crystal structure ${\bf C}$ can be represented by a unit cell with periodic patterns. We focus on 3D crystal structures. More specifically, a unit cell is the minimal repeatable structure of a given crystal and can be described by two matrices {\bf A} and {\bf P}. Matrix ${\bf P}=[p_1,p_2,...,p_n]^T\in \mathbb{R}^{n\times 3}$ is the coordinate matrix where $p_i\in \mathbb{R}^3$ is the 3D Cartesian coordinate of the $i$-th atom in the unit cell. Matrix ${\bf A}=[a_1,a_2,...,a_n]\in \mathbb{Z}^{n}$ is the atom type matrix where $a_i$ is the atomic type of the $i$-th atom. For the periodic patterns, a lattice matrix ${\bf L}=[l_1,l_2,l_3]^T\in \mathbb{R}^{3\times 3}$ is used to describe how the unit cell repeats itself in three directions $l_1,l_2,l_3$ to form the whole crystal structure. Given the crystal representation ${\bf C}=({\bf A,P,L})$, the infinite crystal structure can be represented as
\begin{equation}
		{\bf C}=\{(\hat{p_i},\hat{a_i})|\hat{a_i}=a_i,\hat{p_i}=p_i+k_1l_1+k_2l_2+k_3l_3,k_1,k_2,k_3\in\mathbb{Z},1\leqslant i\leqslant n\}
\end{equation}
where $\hat{p_i}$ corresponds to all periodic translations of $p_i$, and all such translations retain the atom type $a_i$. The goal of the crystal property prediction task is to learn a function $f:{\bf C}\rightarrow \mathbb{R}$ that maps a crystal structure to its property value.

\subsection{Mathematical Properties of Quotient Complex}
The proposed quotient complex characterizes both periodic information and higher-order interactions. There are different ways to define equivalence relations, which result in entirely different quotient complexes. Mathematically, for a topological space $X$, one can define an equivalence relation $\sim$ on $X$ and obtain the set of equivalence classes $X/\sim = \{ [x] : x \in X \}$, along with the canonical quotient map $q : X \rightarrow X/\sim$, where $x \mapsto [x]$. The quotient topology on $X/\sim$ is the finest topology that makes the map $q$ continuous. Throughout this section, we use $\sim_{period}$ to denote the periodic equivalence relation on the atoms of the crystal structure. When transitioning to the representation in crystal simplicial complexes, this relation is interpreted as an equivalence relation on the vertex set of the simplicial complex.

\begin{definition}\label{definition:adjunction}
	Let $X$ be a topological space and $A$ a subset of $X$. For a continuous map $f:A\rightarrow Y$, the adjunction space of $f$ denoted by $X\cup_fY$, is defined as the quotient space $X\sqcup Y/\sim$, where $X\sqcup Y$ is the disjoint union of $X$ and $Y$, $\sim$ is the equivalence relation such that $a\sim f(a)$ for all $a\in A\subset X$.
\end{definition}

\begin{lemma}\label{lemma:quotient1}
	Given a crystal simplicial complex representation $\mathcal{K}$ with vertex set $A$, let $Z$ be the set of equivalent classes of $A$ under periodic equivalent relation $\sim_{period}$. There is a natural projection $f:A\rightarrow Z$. Then the adjunction space $X\cup_fY$ is exactly the quotient complex derived from $\mathcal{K}$ by identifying equivalent atoms up to $\sim_{period}$. We denote $X\cup_fY$ by $\overline{\mathcal{K}}$.
\end{lemma}
Lemma \ref{lemma:quotient1} elucidates that the quotient complex $\overline{\mathcal{K}}$, obtained from the crystal simplicial complex $\mathcal{K}$, can be viewed as an adjunction space. Note that $\overline{\mathcal{K}}$ is usually not a simplicial complex. Subsequently, we provide a construction of a simplicial complex that is homotopy equivalent to $\overline{\mathcal{K}}$, thereby enabling the application of computational tools designed for simplicial complexes to the quotient complex.

\begin{lemma}\label{lemma:quotient2}
	Given a crystal simplicial complex representation $\mathcal{K}$ with vertex set $A$, we build a simplicial complex $S$ as follows:
	\begin{enumerate}
		\item Add all elements of $A$ to $S$.
		\item For every pair of elements $(a_1,a_2)$ in $A$, if they satisfy  $a_1\sim_{period}a_2$, add a new vertex $a_{12}$ and two edges $(a_1,a_{12})$ and $(a_2,a_{12})$ to $S$.
	\end{enumerate}
	We have the inclusion map $i:A\rightarrow S$, and we denote the adjunction space $\mathcal{K}\cup_iS$ by $\widetilde{\mathcal{K}}$.
\end{lemma}

It is noteworthy that $\widetilde{\mathcal{K}}$ remains a simplicial complex. Subsequently, we proceed to demonstrate that there exists a homotopy equivalence between $\widetilde{\mathcal{K}}$ and $\overline{\mathcal{K}}$.

\begin{theorem}\label{theorem:homotopy-equivalent}
	For a crystal simplicial complex representation $\mathcal{K}$, the quotient complex $\overline{\mathcal{K}}$ in lemma \ref{lemma:quotient1} and the quotient complex $\widetilde{\mathcal{K}}$ in lemma \ref{lemma:quotient2} are homotopy equivalent. Further, $\mathcal{K}$ is a subcomplex of $\widetilde{\mathcal{K}}$.
\end{theorem}
The homotopy equivalence in theorem \ref{theorem:homotopy-equivalent} means $\widetilde{\mathcal{K}}$ and $\overline{\mathcal{K}}$ have the same homotopy properties, like homology groups and homotopy groups.

\begin{theorem}\label{theorem:periodic}
	Given a crystal simplicial complex representation $\mathcal{K}$, let $\widetilde{\mathcal{K}}$ be the quotient complex in lemma \ref{lemma:quotient2}, then the inclusion map $\mathcal{K}\hookrightarrow\widetilde{\mathcal{K}}$ induces a homomorphism denoted as $\theta_\bullet:H_\bullet(\mathcal{K})\rightarrow H_\bullet(\widetilde{\mathcal{K}})$. We have
	\begin{itemize}
		\item[(a)] $\theta_0: H_0(\mathcal{K}) \rightarrow H_0(\widetilde{\mathcal{K}})$ is onto;
		\item[(b)] $\theta_1: H_1(\mathcal{K}) \rightarrow H_1(\widetilde{\mathcal{K}})$ is one-to-one;
		\item[(c)] $\theta_q: H_q(\mathcal{K}) \rightarrow H_q(\widetilde{\mathcal{K}})$ is an isomorphism for $q > 1$.
	\end{itemize}
\end{theorem}
Theorem \ref{theorem:periodic} demonstrates how the quotient complex captures the periodic patterns inherent in crystal structures. Specifically, subpart $(a)$ of the theorem states that the quotient complex $\widetilde{\mathcal{K}}$ retains the topological information that is invariant under periodic equivalence in the crystal simplicial complex $\mathcal{K}$, while eliminating redundant topological information. Conversely, subpart $(b)$ indicates that the quotient complex $\widetilde{\mathcal{K}}$ incorporates additional new periodic information compared to the original simplicial complex representation $\mathcal{K}$. The conclusion in subpart $(c)$ follows from the fact that the quotient operation acts specifically on the vertices.

Furthermore, we apply persistent homology (PH), a crucial framework in TDA, to elucidate how the QC structure captures periodicity within the material. Additionally, PH provides extra features and information, shaping more intricate structures within the material neural network. Integrating these PH features into the proposed QCformer to enhance its performance is one of the key areas of our future research. All theoretical results are detailed in Supplementary materials \ref{Appendix: Mathematical Analysis on Quotient Complexes}.

\subsection{QCformer for Perovskite Data Analysis}
\subsubsection{Model Architecture}
\label{methods:model}
As illustrated in Figure \ref{fig:model}, the model first constructs a two-dimensional quotient complex from the crystal structure, where vertices correspond to atoms in a unit cell, edges represent pairwise interactions between atoms, and triangles capture three-body interactions among three atoms. Feature information is embedded in the vertices, edges, and triangles, and these simplex features are updated through a message-passing scheme. Importantly, features on simplices of different dimensions are updated in distinct modules. First, 0-dimensional features are updated by five layers, receiving messages only from 1-dimensional features. Then, higher-order messages flow from 2-dimensional simplices to 0-dimensional simplices in a simultaneous process across two layers. Unlike conventional approaches \cite{papillon2023architectures,hajijtopological}, where features across dimensions are updated simultaneously using an identical number of layers, our approach prioritizes a broader receptive field for 0-dimensional simplices while preserving higher-order information to prevent its loss through successive message-passing layers. As illustrated in Figure \ref*{fig:model}, the $0$-simplices (vertices) are first updated by five Node Transformer layers that take $0$-simplices and $1$-simplices as inputs. Subsequently, the 0-simplices are updated by two simultaneous Edge-Node Transformer layers, where Edge Transformer layers process $1$-simplices and $2$-simplices as inputs. In this message-passing scheme, the attention mechanism aggregates messages from an $n$-simplex's neighbors ($n$-simplices) and cofaces ($(n+1)$-simplices).  Detailed simplex features (vertex, edge, and triangle) and model hyperparameters can be found in Supplementary materials \ref{Appendix: Model Implementation Details}.

\subsubsection{Higher Order Analysis}
In our QCformer model, we represent the crystal structure using a two-dimensional simplicial complex, which includes edge features to encode pairwise interactions and triangle features to capture three-body interactions. To better understand the significance of higher-order interactions, we conduct an ablation study by removing the triangle features from the representation. The resulting model, which reduces to a quotient graph neural network, is referred to as 1D-QCformer.
The performance comparison between 1D-QCformer and QCformer is presented in Table
\ref{table:mp-result-graph} and Table  \ref{table:jarvis-result-graph} for the Materials Project and JARVIS datasets, respectively.
\begin{table}[h]
	\caption{\textbf{Model performance (MAE) comparison between QCformer and 1D-QCformer on the Materials Project dataset.} The best results are shown in {\bf bold}. The units are eV/atom, eV, log(GPa), log(GPa) for Formation Energy, Band Gap, Bulk Moduli and Shear Moduli, respectively.}
	\label{table:mp-result-graph}
	\vskip 0.15in
	\begin{center}
				\begin{tabular}{l|cc}
					\hline
					Task&1D-QCformer &  {\bf QCformer}\\
					\hline
					Formation Energy& 0.019 & {\bf 0.018}\\
					\hline
					Band Gap &   0.195& {\bf 0.191}\\
					\hline
					Bulk Moduli &0.037 &{\bf 0.033}\\
					\hline
					Shear Moduli &0.065 &{\bf 0.059}\\
					\hline
				\end{tabular}
	\end{center}
	\vskip -0.1in
\end{table}
\begin{table}[h]
	\caption{\textbf{Model performance (MAE) comparison between  QCformer and 1D-QCformer on the JARVIS dataset.} The best results are shown in {\bf bold}. The units are eV/atom, eV/atom, eV, eV, eV for Formation Energy, Total Energy, Bandgap(OPT), Bandgap(MBJ) and ${\text E}_{\text{hull}}$, respectively.}
	\label{table:jarvis-result-graph}
	\vskip 0.15in
	\begin{center}
				\begin{tabular}{l|cc}
					\hline
					Task&1D-QCformer &  {\bf QCformer}\\
					\hline
					Formation Energy& 0.0298 & {\bf 0.0290}\\
					\hline
					Total Energy &   0.030& {\bf 0.029}\\
					\hline
					Bandgap(OPT) &0.117 &{\bf 0.113}\\
					\hline
					Bandgap(MBJ) &0.27 &{\bf 0.24}\\
					\hline
					${\text E}_{\text{hull}}$  &  0.059 & {\bf 0.053} \\
					\hline
				\end{tabular}
	\end{center}
	\vskip -0.1in
\end{table}
It can be seen that QCformer consistently outperforms 1D-QCformer across all prediction tasks on both datasets. Specifically, for the Materials Project dataset, QCformer achieves MAE improvements of 5.26\%, 2.05\%, 10.81\% and 9.23\% for Formation Energy, Band Gap, Bulk Moduli and Shear Moduli, respectively. Similarly, for the JARVIS dataset, QCformer outperforms 1D-QCformer with MAE improvements of 2.68\%, 3.33\%, 3.42\%, 11.11\% and 10.17\% for Formation Energy, Total Energy, Bandgap(OPT), Bandgap(MBJ) and ${\text E}_{\text{hull}}$, respectively. These results underscore the critical role of higher-order interactions in effectively representing material properties.

\subsubsection{QCformer for 2D Perovskites Design}
\label{methods:new-prediction}
In the study of new 2D perovskites, our model yields a relatively high MAE of 0.3849 eV across the predictions for five newly identified compounds. Notably, for materials 309, 315, and 363, the predicted bandgaps closely match the DFT-calculated values, with absolute errors of 0.1104, 0.2332, and 0.0096 eV, respectively. However, for material 452, the absolute prediction error is significantly larger at 0.6818 eV. This discrepancy may stem from the inherent limitations of DFT methods in bandgap calculations. As shown in Table S2, different DFT functionals yield significantly varying bandgap results. In this study, we employed the r2SCAN exchange-correlation functional for bandgap predictions. Typically, r2SCAN bandgap values tend to be higher than those obtained using the GGA-PBE functional. The MAE between these two DFT methods is 0.4297 eV, which is comparable to the 0.3849 eV achieved by our QCformer model. These findings emphasize that computational biases can exist among different DFT methodologies. Considering these biases, our model demonstrates its ability to provide reliable estimates of key electronic properties for novel 2D perovskite materials.


\clearpage 

%
\bibliography{refs} 
\bibliographystyle{sciencemag}

%
%
%
%
%
%


\section*{Acknowledgments}
The authors thank Xian Wei, Zheng Wan and Xiangxiang Shen from East China Normal University for the discussion of Sformer code. 
\paragraph*{Funding:}
This work was supported in part by the Singapore Ministry of
Education Academic Research Fund Tier 1 grant RG16/23,
Tier 2 grants MOE-T2EP20120-0010, MOE-T2EP20221-
0003 and MOE-T2EP50123-0001, as well as the National
Research Foundation (NRF), Singapore, under its Competitive
Research Program (CRP) (NRF-CRP25-2020-0004).
\paragraph*{Author contributions:}
K.X. conceived this idea and led the project. C.-S.H. analyzed the theory. X.L. and X.Y. implemented the models and performed the experiments. All authors analyzed the data, discussed the results, and contributed to the manuscript.
\paragraph*{Competing interests:}
There are no competing interests to declare.
\paragraph*{Data and materials availability:}
The Materials Project and JARVIS dataset are accessible through the JARVIS-Tools python package. 
The HOIP dataset can be accessed via Dryad
Digital Repository: \url{http://dx.doi.org/10.5061/dryad.gq3rg}. 
The HOIP2D dataset is available on the official website of the NMSE database: \url{http://www.pdb.nmse-lab.ru/}.
Code for reproducing our results is available online from the repository: 
\url{https://github.com/xinyu030/QCformer}.




\subsection*{Supplementary materials}
Mathematical Analysis on Quotient Complexes\\
Quotient Complex Construction and Feature Generation\\
Data and Experiments\\
Figs. S1 to S5\\
Table S1 to S2\\
References \textit{(70-\arabic{enumiv})}\\ 


\newpage


\renewcommand{\thefigure}{S\arabic{figure}}
\renewcommand{\thetable}{S\arabic{table}}
\renewcommand{\theequation}{S\arabic{equation}}
\renewcommand{\thepage}{S\arabic{page}}
\setcounter{figure}{0}
\setcounter{table}{0}
\setcounter{equation}{0}
\setcounter{page}{1} 


\begin{center}
\section*{Supplementary Materials for\\ \scititle}

Xinyu You,
Xiang Liu,
Chuan-Shen Hu,
Kelin Xia$^\ast$,
Tze Chien Sum$^\ast$\\ 
\small$^\ast$Corresponding author. Email: xiakelin@ntu.edu.sg; tzechien@ntu.edu.sg.
\end{center}

\subsubsection*{This PDF file includes:}
Mathematical Analysis on Quotient Complexes\\
Quotient Complex Construction and Feature Generation\\
Data and Experiments\\
Figs. S1 to S5\\
Table S1 to S2\\


\newpage


\renewcommand{\thesection}{S\arabic{section}}
\renewcommand{\thesubsection}{S\arabic{section}.\arabic{subsection}}
\renewcommand{\thesubsubsection}{S\arabic{section}.\arabic{subsection}.\arabic{subsubsection}}
\setcounter{section}{0} 
\setcounter{subsection}{0}
\setcounter{subsubsection}{0}

\section{Mathematical Analysis on Quotient Complexes}
\label{Appendix: Mathematical Analysis on Quotient Complexes}

In this section, we provide the mathematical foundation for constructing the quotient complex (QC) from crystal data, a pivotal element in the QCformer architecture. Specifically, we outline essential mathematical concepts in Section \ref{Appendix: Simplicial Complexes}, covering simplicial complexes, ordered simplicial complexes, and clique complexes.

In Section \ref{Appendix: Quotient Complexes}, we present the quotient operation applied to simplicial complexes for the generation of QCs. Beginning with arbitrary topological spaces, we introduce equivalence relations and provide examples of QCs derived from simplicial complexes. Additionally, we analyze the topology of QCs using homology theory, illustrating how the quotient operation impacts the original topology established from the atomic system of the material.


\subsection{Simplicial Complexes}
\label{Appendix: Simplicial Complexes}

In this section, we introduce fundamental geometric objects employed in this work, encompassing simplicial complexes, ordered simplicial complexes, and clique complexes—topics prevalent in algebraic topology~\cite{munkres2018elements} and widely applied to simplicial neural networks~\cite{papillon2023architectures,hajijtopological}. While these concepts have been explored in various literature, the notations employed often differ. Consequently, we provide a concise recapitulation of these concepts, unifying the notations for clarity in this paper.

\paragraph{Simplicial Complexes}
An \textit{abstract simplicial complex} $K$ over a finite vertex set $V$ is a collection of non-empty finite subsets of $V$ such that if $\sigma \in K$ and $\tau \subseteq \sigma$, then $\tau \in K$. Any element $\sigma \in K$ with $n+1$ vertices is termed an $n$-simplex, where $n$ is also known as the \textit{dimension} of $\sigma$. The dimension of an abstract simplicial complex is defined as the highest dimension among its simplices.

An abstract simplicial complex serves as a blueprint for real simplices embedded in Euclidean space, with a particular emphasis on the face relations between simplices. Specifically, any subset $\tau$ of $\sigma$ is referred to as a \textit{face} of $\sigma$, and $\sigma$ is denoted as a \textit{coface} of $\tau$.

Geometrically, $0$-simplices correspond to vertices, 1-simplices to edges, 2-simplices to triangles, 3-simplices to tetrahedra, and so forth for higher-dimensional simplices. The embedding of an abstract simplicial complex into Euclidean space is referred to as the \textit{geometric realization}~\cite{munkres2018elements}. We use $\mathcal{K}$ to denote the embedded space and simply refer to it as a \textit{simplicial complex}. In particular, since $V$ is a finite set, the embedded $\mathcal{K}$ is a compact space~\cite{munkres2018elements}.

\paragraph{Ordered Simplicial Complexes}

An \textit{ordered simplicial complex} $K$ over a finite vertex set $V$ is an abstract simplicial complex with orderings on its simplices. Mathematically, each $n$-simplex in $K$ is represented as an ordered sequence $\sigma = (v_0, v_1, \ldots, v_n)$ of $n+1$ vertices. Furthermore, every simplex represented as a subsequence of $\sigma$ is also required to be an element in $K$~\cite{lutgehetmann2020computing}.

In particular, for any $n$-simplex $\sigma = (v_0, v_1, \ldots, v_n)$ and $i \in \{ 0, 1, ..., n \}$, the ordered subset $\sigma^i := (v_0, \ldots, \widehat{v_i}, \ldots, v_n) := (v_0, \ldots, v_{i-1}, v_{i+1}, \ldots, v_n)$, obtained by removing the $i$-th vertex from $\sigma$, is called the $i$-th face of $\sigma$. Similar to an abstract simplicial complex, $\sigma$ is a \textit{coface} of the simplices $\sigma^i$. Two $n$-simplices $\sigma$ and $\tau$ are called \textit{upper adjacent} if they share a common coface, denoted by $\sigma \frown \tau$. 


In this context, an undirected graph can be generalized to a simplicial complex, and a directed graph can be generalized to an ordered simplicial complex. Specifically, both of them are abstract and ordered simplicial complexes of dimension at most $1$. Beyond the interactions encoded by edges, higher-dimensional simplices are capable of capturing more complex and intricate interactions involving more than two nodes or vertices.

\paragraph{Clique Complex of a Directed Graph}  Given a directed graph $G=(V,E)$ with vertex set $V$ and arrow set $E$. A \textit{clique} $\sigma$ of $G$ is a subsequence $\sigma=(v_{i_0},v_{i_1},v_{i_2},...,v_{i_n})$ of $V$ such that $(v_{i_j},v_{i_k})\in E$ for all $j<k$. All the cliques of $G$ form an ordered simplicial complex, which is called the \textit{clique complex} of the directed graph $G$. For example, if the directed graph $G=(V,E)$ consists of three vertices $1,2$ and $3$ and three arrows $(1,2),(1,3),(2,3)$, then the associated clique complex is
\begin{equation*}
	\{ (1), (2), (3), (1,2), (1,3), (2,3), (1,2,3) \},
\end{equation*}
which contains an ordered $2$-simplex $(1,2,3)$. However, if the arrows set is defined by $\{ (1,2),(2,3),(3,1) \}$, then the induced clique complex contains no ordered $2$-simplices.

\subsection{Quotient Complexes}
\label{Appendix: Quotient Complexes}

In topology, the quotient operation is a fundamental tool for generating new topological spaces from a given topological space. Mathematically, for a topological space $X$, one can define an equivalence relation $\sim$ on $X$ and obtain the set of equivalence classes $X/\sim = \{ [x] : x \in X \}$ along with the canonical quotient map $q : X \rightarrow X/\sim$, where $x \mapsto [x]$. The quotient topology on $X/\sim$ is the finest topology that makes the map $q$ continuous.

In this paper, our focus lies on simplicial complexes $\mathcal{K}$ embedded in the $D$-dimensional Euclidean space $\mathbb{R}^D$. We define equivalence relations $\sim_V$ on their vertex set $V$ and examine the induced quotient complex $\mathcal{K}/\sim_V$.
\begin{figure}[ht]
	\centering
	\includegraphics[width=0.8\textwidth]{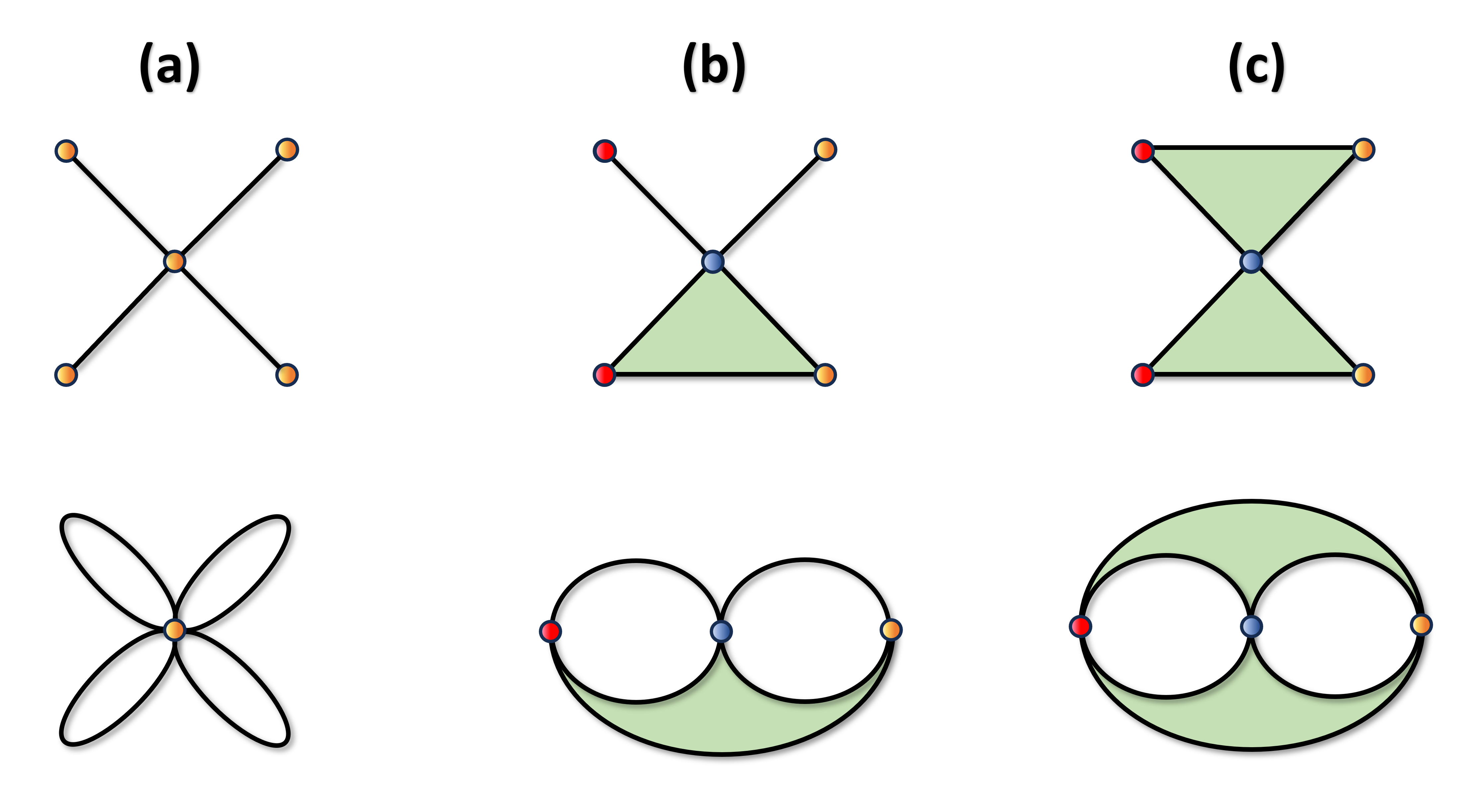}
	\caption{\textbf{Illustration of simplicial complexes and their corresponding quotient complexes.} In these examples, equivalent vertices are indicated by the same color. The quotient operation merges equivalent vertices in the simplicial complexes, resulting in the derived quotient complexes.}
	\label{fig:quotient-complex2}
\end{figure}

Figure \ref{fig:quotient-complex2} illustrates examples of quotient complexes (in the second row) obtained from simplicial complexes (in the first row) by merging equivalent vertices, representing $0$-dimensional simplices. In these examples, equivalent vertices are colored with the same color. By merging equivalent vertices, the quotient complexes are obtained and depicted in the second row. For instance, in {\bf (a)}, the simplicial complex consists of 5 vertices and 4 edges. By identifying all the vertices as the same vertex, the induced quotient complex has only one vertex and four self-loops. On the other hand, the quotient complex in {\bf (b)} includes 5 vertices, 5 edges, and 1 triangle. The 5 vertices are classified into three equivalence classes, colored in red, blue, and orange, respectively. By gluing equivalent vertices separately, the quotient complex is obtained. Finally, the simplicial complex in {\bf (c)} comprises 5 vertices, 6 edges, and 2 triangles. Using the same vertex classification as in {\bf (b)}, the induced quotient complex is derived. This quotient complex also includes two "torsion" triangles, sharing the same vertices.

\paragraph{Adjunction Spaces} Let $X$ denote a topological space and $A$ a subset of $X$. For a continuous map $f: A \rightarrow Y$, one can define the \textit{adjunction space} denoted as $X \cup_f Y$. This space is constructed as the quotient space $X \sqcup Y / \sim$, where $X \sqcup Y$ denotes the disjoint union space of $X$ and $Y$, and $\sim$ represents the least equivalence relation generated by $a \sim f(a)$ for all $a \in A \subseteq X$. In algebraic topology, this quotient operation forms the foundation for constructing cell complexes and serves as a powerful tool for exploring homology theory within the realm of topological manifolds~\cite{vick2012homology}.

As a specific instance of a quotient space, certain special adjunction spaces $X \cup_f Y$ can be realized through a more straightforward equivalence relation applied to the space $X$. This insight facilitates a convenient technique for identifying and gluing specific points within $X$ and proves to be a valuable tool for exploring quotient complexes. For example, the following proposition offers an alternative way to represent the quotient space $X/\sim_A$.

\begin{proposition}[\cite{brown2006topology}]
	\label{Proposition: Quotient and Adjunction Spaces}
	Let $X$ and $Y$ be topological spaces, $A$ a subspace of $X$, and $f: A \rightarrow Y$ a continuous map. Then there is a continuous map $\phi: X/\sim_A \rightarrow X \cup_f Y$ that maps every equivalence class $[x]$ in $X/\sim_A$ to the equivalence class $[x]$ in $X \cup_f Y$. Furthermore, if $f$ is onto and $Y$ is discrete, then $\phi$ is a homeomorphism.
\end{proposition}

Specifically, let $\mathcal{K}$ be the realization of $K$ over a finite vertex set $V$, $\sim_V$ an equivalence relation on $V$, and $Z$ a discrete topological space. Suppose there is a surjective function $f: V \rightarrow Z$ such that $f(v) = f(w)$ if $v \sim_V w$. Notably, $Z$ is finite since there are only finitely many equivalence classes in $V$. Then the induced quotient complex $\overline{\mathcal{K}}$ is homeomorphic to the adjunction space $\mathcal{K} \cup_f Z$. This identification proves useful for the subsequent exploration of the topological structure of the quotient complex.
\begin{figure}[ht]
	\centering
	\includegraphics[width=1\textwidth]{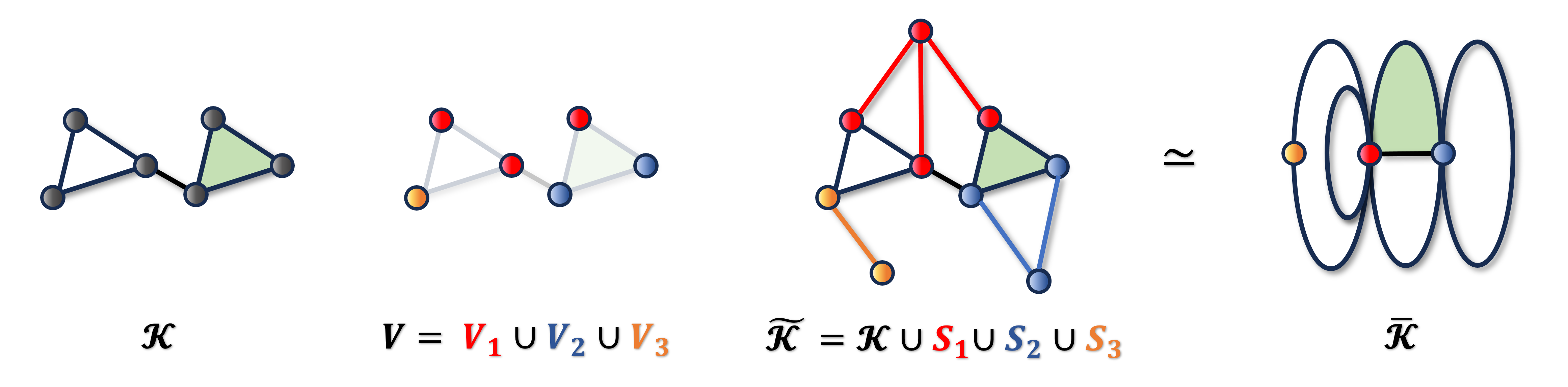}
	\caption{\textbf{An illustrative example depicting the construction of the simplicial complex $\widetilde{\mathcal{K}}$, which is homotopy equivalent to the quotient complex $\overline{\mathcal{K}}$.} In this example, $\mathcal{K}$ denotes a simplicial complex with 6 vertices., 7 edges, and 1 triangle. The vertex set $V$ is divided into three groups: $V_1$, $V_2$, and $V_3$. By adding star-shaped sets $S_1$, $S_2$, and $S_3$ to the simplicial complex $\mathcal{K}$, we obtain the simplicial complex $\widetilde{\mathcal{K}}$, which is homotopy equivalent to the quotient complex $\overline{\mathcal{K}}$.}
	\label{fig:homotopy_equivalence}
\end{figure}

\paragraph{Homotopy Equivalent Construction} In our earlier discussions, we introduced the quotient operation applied to a finite simplicial complex. To be specific, we delineated a set of vertices, denoted as $V$, within the complex $K$. As shown in Figure \ref{fig:quotient-complex2}, for the geometric realization $\mathcal{K}$, we create the quotient space $\overline{\mathcal{K}}$ by merging equivalent points in $V$ separately. Additionally, we propose a method for constructing a simplicial complex, referred to as $\widetilde{\mathcal{K}}$, which is homotopy equivalent to $\overline{\mathcal{K}}$.

To elucidate the homotopy equivalence relation between the constructed $\widetilde{\mathcal{K}}$ and the quotient complex $\overline{\mathcal{K}}$, we refer to the following well-known theorem in homotopy theory.

\begin{theorem}[\cite{brown2006topology}]
	\label{homotopy theorem in Brown's book}
	Suppose the following diagram of topological spaces and continuous maps
	\begin{equation*}
		\xymatrix@+1.0em{
			& Y
			\ar[d]_{\varphi_Y}
			& A
			\ar[l]_{f}
			\ar@{^{(}->}[r]^{i}
			\ar[d]^{\varphi_A}
			& X
			\ar[d]^{\varphi_X}
			\\
			& Y'
			& A'
			\ar[l]^{f'}
			\ar@{^{(}->}[r]_{i'}
			& X'
		}
	\end{equation*}
	commutes, where $\varphi_X, \varphi_A, \varphi_Y$ are homotopy equivalences, and the inclusions $i, i'$ are closed cofibrations. Then the map $\varphi: X \cup_f Y \longrightarrow X' \cup_{f'} Y'$ induced by the maps $\varphi_X, \varphi_A, \varphi_Y$ is a homotopy equivalence.
\end{theorem}

The mathematical formulation of the space $\widetilde{\mathcal{K}}$ is established as follows. Let $V_1, ..., V_k$ be disjoint subsets of the vertex set $V$ of $\mathcal{K}$. Let $z_1, z_2, ..., z_k$ in $\mathbb{R}^{D+k}$ such that $z_{j + 1} \in \mathbb{R}^{D+j+1} \setminus \mathbb{R}^{D+j}$, where we identify $\mathbb{R}^{D+j}$ as a subspace of $\mathbb{R}^{D+j+1}$ via the canonical embedding map $(x_1, ..., x_{D+j}) \mapsto (x_1, ..., x_{D+j}, 0)$. We set $V = \bigsqcup_{j = 1}^k V_j$ and $Z := \{ z_1, z_2, ..., z_k \}$, which is a discrete space. By Proposition \ref{Proposition: Quotient and Adjunction Spaces}, the desired space $\overline{\mathcal{K}}$ that emerges points in $V_i$ to a single point can be identified as the adjunction space $\mathcal{K} \cup_f Z = \mathcal{K} \cup_f \{ z_1, z_2, ..., z_k \}$, where $f: V \rightarrow Z$ is the local constant function with $f(V_j) = \{ z_j \}$. Next, for each $j \in \{ 1, 2, ..., k \}$ we define a star-shaped set $S_j$ as the union of all line segments that link $z_j$ to points in $V_j$. That is,
\begin{equation*}
	S_j = \bigcup_{x \in V_j} [z_j, x] = \bigcup_{x \in V_j} \{ \lambda z_j + (1-\lambda) x \ | \ \lambda \in [0,1] \}.
\end{equation*}
In particular, $S_i \cap S_j = \emptyset$ for $i \neq j \in \{ 1,2, ..., k\}$ by the assumption of $z_1, z_2, ..., z_k$. Set $S = \bigsqcup_{j = 1}^k S_j$ and define $g: V \rightarrow S$ such that $g|_{V_j}$ equals the inclusion map $V_j \hookrightarrow S$. Finally, we set $\widetilde{\mathcal{K}} = \mathcal{K} \cup S = \mathcal{K} \cup_g S$. Then the diagram
\begin{equation*}
	\xymatrix@+1.0em{
		& S
		\ar[d]_{r}
		& V
		\ar[l]_{g}
		\ar@{^{(}->}[r]^{i}
		\ar[d]^{{\rm id}_V}
		& \mathcal{K}
		\ar[d]^{{\rm id}_\mathcal{K}}
		\\
		& Z
		& V
		\ar[l]^{f}
		\ar@{^{(}->}[r]_{i}
		& \mathcal{K}
	}
\end{equation*}
commutes, where $r: S \rightarrow Z$ is defined such that $r|_{S_j}$ equals the constant map $S_j \rightarrow \{ z_j \} \hookrightarrow Z$. Since the continuous maps $r|_{S_j}$ can be shrunk synchronously, $r$ is a homotopy equivalence. Because any inclusion map of a subcomplex of a simplicial complex is a closed cofibration, $i$ is a closed cofibration (\cite{brown2006topology}, page 281). By Theorem \ref{homotopy theorem in Brown's book}, the following corollary follows.

\begin{corollary}
	\label{Main corollary 1}
	Let $K$, $V = \bigsqcup_{j = 1}^k V_j$, $\mathcal{K}$,  $\overline{\mathcal{K}}$, and $\widetilde{\mathcal{K}}$ be defined as above. Then $\overline{\mathcal{K}}$ and $\widetilde{\mathcal{K}}$ are homotopy quivalent. In particular, the homology of $\overline{\mathcal{K}}$ and $\widetilde{\mathcal{K}}$ are isomorphic.
\end{corollary}

In other words, there is a homotopy equivalence $\widehat{r}$ from $\mathcal{K} \cup_g S = \mathcal{K} \cup S = \widetilde{\mathcal{K}}$ to $\mathcal{K} \cup_f Z = \overline{\mathcal{K}}$, which is induced by the homotopy equivalence $r: S \rightarrow Z$.

Figure \ref{fig:homotopy_equivalence} visually demonstrates the result of Corollary \ref{Main corollary 1} through illustrative graphics.  Leveraging the simplicial complex $\mathcal{K}$ and the vertex set $V = \bigsqcup_{j = 1}^k V_j$, we construct the simplicial complex $\widetilde{\mathcal{K}}$ as the union $\mathcal{K} \cup S$, with $S = S_1 \cup S_2 \cup S_3$ denoting the union of star-shaped sets as illustrated in Figure \ref{fig:homotopy_equivalence}. Following the implications of Corollary \ref{Main corollary 1}, both the simplicial complex $\widetilde{\mathcal{K}}$ and the quotient space $\overline{\mathcal{K}}$ share the same homotopy type. Notably, $\widetilde{\mathcal{K}}$ and $\overline{\mathcal{K}}$ exhibit identical homology groups.

In summary, we have devised a method for representing $\widetilde{\mathcal{K}}$ by $\overline{\mathcal{K}}$ up to homotopy equivalence. Importantly, we establish the inclusion relationship $\mathcal{K} \subseteq \widetilde{\mathcal{K}}$. In the following sections, we will see that this representation offers a combinatorial perspective to the study of persistent homology within quotient complexes, thereby facilitating the analysis and proof of persistent barcode information.




\paragraph{Homology of Quotient Complexes} As a result of the construction presented previously, the space $\widetilde{\mathcal{K}}$ assumes the form of a simplicial complex within $\mathbb{R}^{D+k}$, with $\mathcal{K}$ representing a subcomplex of $\widetilde{\mathcal{K}}$. Consequently, the inclusion map $\mathcal{K} \hookrightarrow \widetilde{\mathcal{K}}$ induces a homomorphism denoted as $\theta_\bullet: H_\bullet(\mathcal{K}) \rightarrow H_\bullet(\widetilde{\mathcal{K}})$. We can now state the following theorem:

\begin{theorem} [\cite{hu2025quotient}]
	\label{Main theorem 1}
	Let $K$, $V = \bigsqcup_{j = 1}^k V_j$, $\mathcal{K}$, and $\widetilde{\mathcal{K}}$ be defined as above. Then
	\begin{itemize}
		\item[\rm (a)] $\theta_0: H_0(\mathcal{K}) \rightarrow H_0(\widetilde{\mathcal{K}})$ is onto;
		\item[\rm (b)] $\theta_1: H_1(\mathcal{K}) \rightarrow H_1(\widetilde{\mathcal{K}})$ is one-to-one;
		\item[\rm (c)] $\theta_q: H_q(\mathcal{K}) \rightarrow H_q(\widetilde{\mathcal{K}})$ is an isomorphism for $q > 1$.
	\end{itemize}
\end{theorem}

Theorem \ref{Main theorem 1}-(a) and (b) elucidates the relations of the homology groups between the quotient complex $\overline{\mathcal{K}} \simeq \widetilde{\mathcal{K}}$ and the original simplicial complex $\mathcal{K}$ in dimensions $0$ and $1$.  On the contrary, Theorem \ref{Main theorem 1}-(c) asserts that the quotient complex $\overline{\mathcal{K}} \simeq \widetilde{\mathcal{K}}$ encapsulates the same homological information as $\mathcal{K}$. The reason is that the gluing operation is only applied to 0-dimensional simplices. Further exploration of the effects of gluing higher-dimensional simplices or subcomplexes is a potential avenue for future research.

\section{Quotient Complex Construction and Feature Generation}
\label{Appendix: Model Implementation Details}
\subsection{Quotient Complex Construction}
\begin{figure}[ht]
	\centering
	\includegraphics[width=0.9\textwidth]{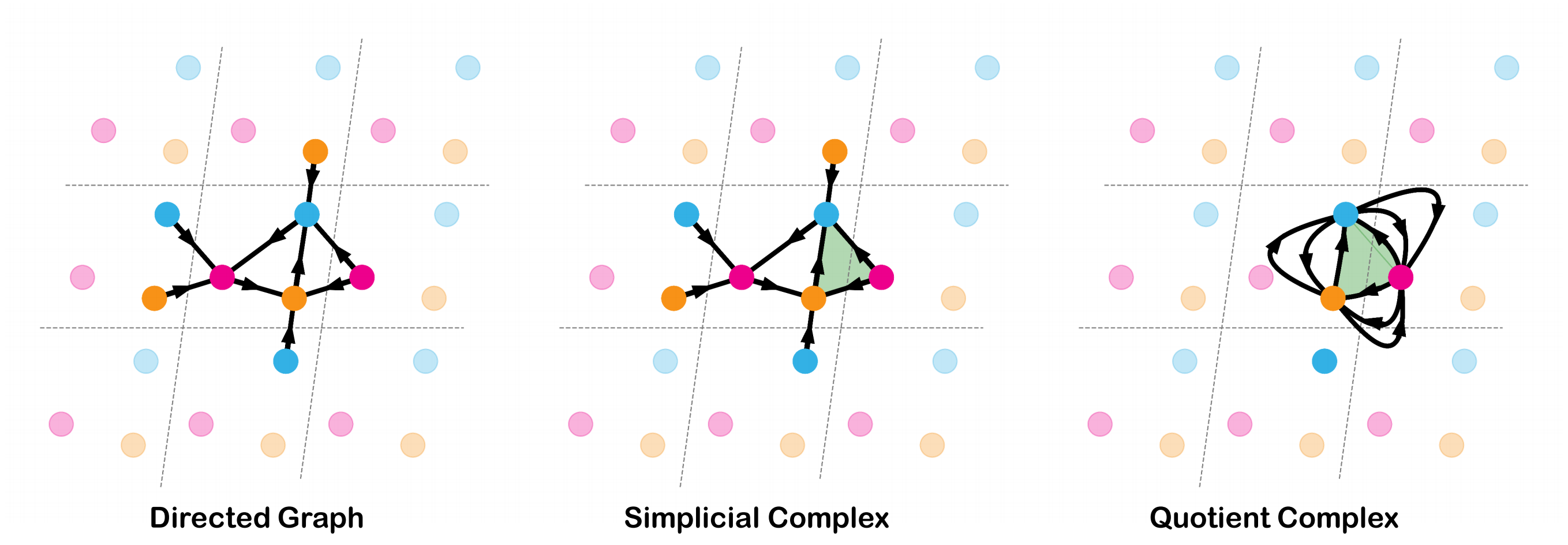}
	\caption{\textbf{Illustration of the crystal quotient complex construction with three nearest neighbors.} First, the atoms within the unit cell are selected as nodes, with directed edges added from the three nearest neighbors to each target node. Next, triangles are formed based on edge directions, constructing the simplicial complex. Finally, nodes of the same color are merged to form the quotient complex.}
	\label{fig:quotient-complex3}
\end{figure}

 For the construction of the crystal quotient complex, the process involves three main steps: first, a directed graph is built from the crystal structure; next, triangles are added to the directed graph to form a simplicial complex; and finally, a quotient operation is applied to the simplicial complex to obtain the final quotient complex. The details of these three steps are as follows:
Firstly, we use $k$-nearest neighbor graph construction for generating the graph. Formally, let the atoms of a unit cell form the vertex set $V$, and denote the nearest $k$ neighbors of a vertex $v$ by $N_{k}(v)$. The directed edge set $E$ is defined as $E=\{(v_i,v_j) \ | \ v_i\in N_{k}(v_j),v_j\in V \}$. The pair $(V,E)$ then forms a directed graph. Secondly, to incorporate triangles, we consider the clique complex of the directed graph. Specifically, for a triangle $(v_i,v_j,v_k)$ where $\{(v_i,v_j),(v_j,v_k),(v_i,v_k)\}\subset E$, we assign a total order to these three edges as $(v_i,v_j)<(v_j,v_k)<(v_i,v_k)$. Finally, for the quotient operation, we define two atoms $a$ and $b$ as equivalent if there exist $k_1,k_2,k_3\in\mathbb{Z}$ such that $a=b+k_1l_1+k_2l_2+k_3l_3$, where $l_1,l_2,l_3$ are the lattice vectors. The quotient complex $C$ is derived by gluing all equivalent atoms together. Figure \ref{fig:quotient-complex3} gives an illustration of this construction for $k=3$.

In our QCformer model, simplex features are updated by aggregating information from their neighbors and cofaces. There are various methods to define the neighbors of a simplex, such as upper adjacent, lower adjacent, parallel adjacent, and so on. In the QCformer model, we define the neighbors of each vertex $\sigma_0$=$(v)$ as $N(\sigma_0)=\{v_j \ | \ (v_j,v)\in E\}$. Similarly, the neighbors of each edge $\sigma_1$=$(v_i,v_j)$ are defined as $N(\sigma_1)=\{\tau_1 \ | \ \sigma_1 \frown \tau_1, \ \tau_1 < \sigma_1\}$
where $\sigma_1 \frown \tau_1$ indicates that $\sigma_1$ and $\tau_1$ are upper adjacent.

\begin{figure}[ht]
	\centering
	\includegraphics[width=0.9\textwidth]{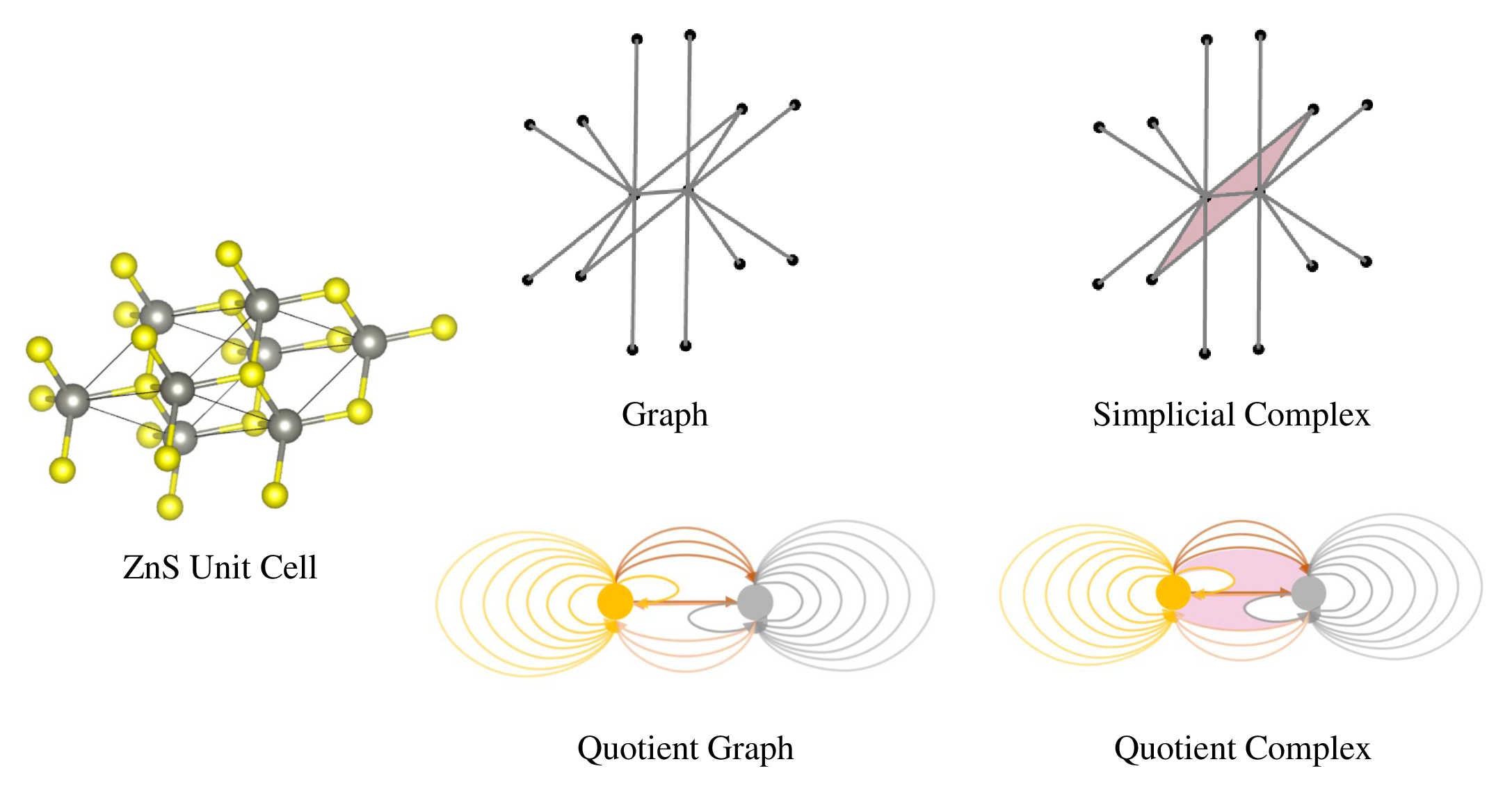}
	\caption{\textbf{Illustration of quotient complex construction on ZnS.}  On the left-hand side, the unit cell structure is visualized using the VESTA program~\cite{momma2011vesta}. On the right-hand side, a graph is constructed by selecting the 12-nearest neighbors of each atom in the unit cell. A clique complex is then formed based on this graph, with 2-simplices highlighted in pink.  Note that, as both the graph and the simplicial complex are three-dimensional objects, the two-dimensional projections shown have overlapping vertices and edges. The quotient graph and quotient complex are constructed by gluing equivarient vertices together. Multiple edges between different vertices are displayed in distinct colors. In particular, the two 2-simplices consist of two normal edges and one self-loop edge.}
	\label{fig:quotient-complex-ZnS}
\end{figure}

Computationally, in the QCformer model, a 12-nearest-neighbor-based crystal graph is constructed, and its clique complex is used to derive the quotient complex through the quotient operation on equivalent atoms. 
A detailed example of the quotient complex construction for ZnS material is provided in figure
\ref{fig:quotient-complex-ZnS}.

\subsection{Simplex Feature Generation}
After generating the quotient complex, we assign features to the simplices. Our QCformer model adopts a two-dimensional simplicial complex strategy, incorporating features for vertices, edges, and triangles. For each vertex, the CGCNN feature \cite{xie2018crystal} is considered, which is a 92-dimensional vector. For each edge, its Euclidean length $d$ is first mapped to $d'=-0.75/d$ and then processed through the radial basis function (RBF), defined as 
${\rm rbf}(d') = {\rm exp}\bigg(\frac{-(d'-c)^2}{\sigma}\bigg)$. The RBF expands $d'$ into a 192-dimensional vector (64$\times$3), where $c$ ranges from -4.0 to 0.0 with a step size of 1/64, and $\sigma$ takes values 0.01, 0.1 and 1. This RBF vector is then concatenated with the 92-dimensional CGCNN feature vectors of the two vertices connected by the edge, resulting in a 376 dimensional edge embedding (192+92$\times$2). For triangles, we use combinations of edge lengths $d_1,d_2,d_3$ to generate additional features, including $d_1\cdot d_2,d_1\cdot d_3,d_2\cdot d_3$, and the squares $d_1\cdot d_1,d_2\cdot d_2,d_3\cdot d_3$. These nine values are passed through the RBF, defined as
${\rm rbf}(a) = {\rm exp}\bigg(\frac{-(a-c)^2}{\sigma}\bigg)$, to expand into a 216-dimensional vector (8$\times$3$\times$9), where $c$ ranges from 0.0 to 5.0 with a step size of 1/8, and $\sigma$ takes values 0.01, 0.1 and 1. This 216-dimensional vector serves as the triangle embedding.

\section{Data and Experiments}

\subsection{Dataset Details}
\label{Appendix: Dataset Details}
\paragraph{Materials Project}
For the Materials Project dataset, four crystal properties are considered, including Formation Energy, Band Gap, Bulk Moduli, and Shear Moduli. Following Matformer \cite{yan2022periodic}, we use the same training, validation, and test set. Specifically, the tasks of Formation Energy and Band Gap share the same training, validation, and test set, including 60,000, 5,000, and 4,239 crystals, respectively; the tasks of Bulk Moduli and Shear Moduli use the same training, validation, and test set, including 4,664, 393 and 393 crystals, respectively. 

\paragraph{JARVIS}
For the JARVIS dataset, five properties are considered, including Formation Energy, Total Energy, Bandgap(OPT), Bandgap(MBJ) and $\text{E}_{\text{hull}}$.
The tasks of Formation Energy, Total Energy, and Bandgap(OPT) use the same training, validation, and test set, including 44,578, 5,572, and 5,572 crystals; the task of Bandgap(MBJ) uses 14,537, 1,817, and 1,817 crystals for training, validation, and test set, respectively.

\paragraph{HOIP}
The HOIP dataset contains 1346 hybrid organic-inorganic perovskites, featuring 16 organic cations, 3 group-IV cations, and 4 halide anions. While both the HSE06 bandgap and the GGA-rPW86 bandgap are reported in the dataset, we only evaluate the HSE06 bandgap in our work as it is closer to the true bandgap of the materials.
A ten-fold cross-validation protocol is used for training and testing.

\paragraph{HOIP2D}
The HOIP2D dataset contains 849 hybrid organic-inorganic two-dimensional (2D) perovskites, containing 180 organic cations, 10 metals and 3 halide anions. Within this collection, 753 compounds are equipped with DFT-based bandgap values, while 238 compounds possess experimental bandgap values. Specially, we only use 716 compounds with DFT-based bandgap values, as supported by the unit cell representation from the pymatgen package. To ensure a fair comparison, we train and test
our model on a randomly selected subset of 624 data points (out of 716 compounds), repeating the
subsampling procedure five times for each run to mitigate randomness. A five-fold cross-validation protocol is used for training and testing.

\subsection{Experimental Setup}
We provide details about our QCformer model. The crystal quotient complex is constructed by considering the 12 nearest neighbors. For each simplex, its feature vector is mapped to a 64-dimensional embedding using an MLP. The message-passing mechanism involves five layers for updating from edges to vertices and two layers for updating from triangles to edges to vertices. Global average pooling is applied to aggregate features from all vertices and edges, followed by an MLP with two hidden layers to predict the crystal property.

The QCformer model is implemented using Pytorch \cite{paszke2019pytorch} and Pytorch Geometric \cite{fey2019fast}. For all prediction tasks, we employ the AdamW \cite{loshchilov2017decoupled} optimizer with a weight decay of 1e-5 and a one-cycle learning rate schedular \cite{smith2019super}. 
Other hyperparameters are listed in Table \ref{tab:training_config}. 
All tests are conducted on an NVIDIA RTX 3090 GPU with 20GB of memory. 

\begin{table}[ht]
\caption{\textbf{Training configurations for Materials Project, JARVIS, HOIP and HOIP2D datasets.} We use mean squared error (MSE) loss and mean absolute error (MAE) loss for inorganic materials (Materials Project and JARVIS) and hybrid organic-inorganic materials (HOIP and HOIP2D), respectively. }
\label{tab:training_config}
\centering
\begin{tabular}{lcccc}
\hline
Dataset        & Batch size & Epochs & Learning rate & Loss function \\ \hline
Materials Project       & 64                  & 1000            & 0.0006                 & MSE                    \\ 
JARVIS                  & 64                  & 1000            & 0.0006                 & MSE                    \\ 
HOIP                    & 64                  & 500             & 0.005                  & MAE                    \\ 
HOIP2D                  & 64                  & 500             & 0.005                  & MAE                    \\ \hline
\end{tabular}
\end{table}

\subsection{Results}

\paragraph{Performance Evaluation Using MAD:MAE} While we already report MAE results for each property in the Materials Project and JARVIS datasets in the main article, the varying units and distinct variances associated with these properties necessitate additional evaluation metrics. To provide a more objective assessment, we include the mean absolute deviation (MAD) and report the MAD:MAE ratio for each property. A model is generally considered a strong predictor if its MAD is at least five times its MAE \cite{choudhary2021atomistic}. In the Materials Project dataset, QCformer achieves MAD:MAE ratios of 49.54, 6.88, 7.89, and 4.82 for Formation Energy, Band Gap, Bulk Modulus, and Shear Modulus, respectively. In the JARVIS dataset, QCformer attains MAD:MAE ratios of 30.08, 58.56, 8.93, 7.08, and 19.60 for Formation Energy, Total Energy, Bandgap (OPT), Bandgap (MBJ), and ${\text E}_{\text{hull}}$, respectively. These results demonstrate QCformer’s strong predictive performance in material property prediction.

\begin{figure}[ht]
\label{appendix: 2d prediction}
	\centering
	\includegraphics[width=0.8\textwidth]{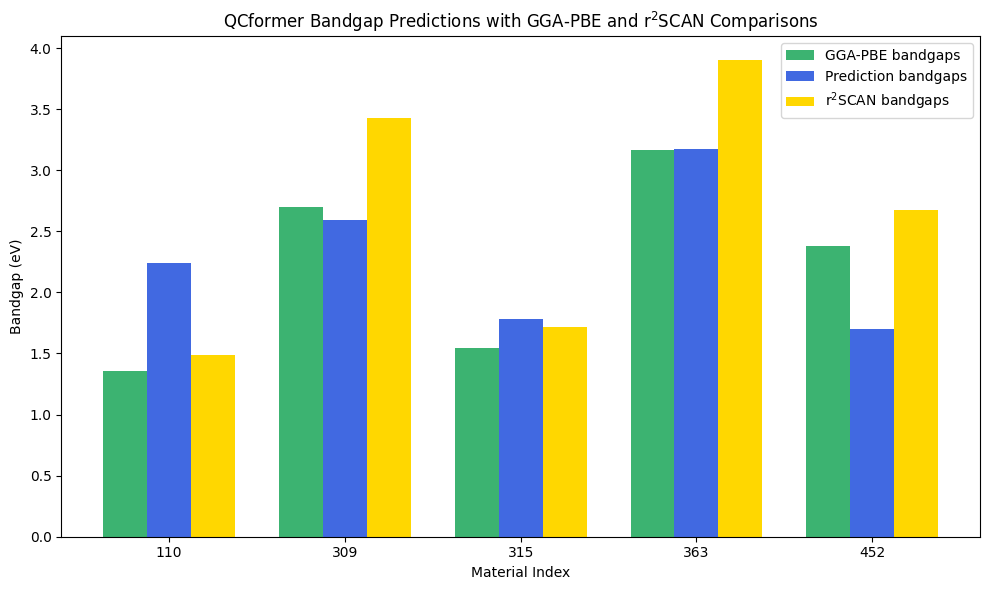}
	\caption{\textbf{Prediction of bandgaps for unlabeled 2D perovskite materials using QCformer.} The figure compares GGA-PBE bandgaps (green) and r$^2$SCAN bandgaps (yellow) with QCformer predictions (blue) for five compounds from the HOIP2D dataset that lacked prior bandgap labels.}
	\label{fig:prediction}
\end{figure}

\paragraph{2D New Perovskites Bandgap Prediction} The bandgaps of five new perovskite materials from the HOIP2D dataset, which lack experimental labels, were predicted using QCformer. The results are compared against GGA-PBE-based DFT bandgaps, as shown in Figure \ref{appendix: 2d prediction}. While QCformer achieves a prediction MAE of 0.3849 eV compared to GGA-PBE, the table \ref{table:new-bandgaps-2} highlights additional comparisons with r$^2$SCAN-based DFT bandgaps. Notably, the MAE between GGA-PBE and r$^2$SCAN bandgaps is 0.4297 eV, which exceeds the QCformer MAE. This comparison underscores the inherent challenges in bandgap prediction due to discrepancies between computational methods.

Figure \ref{appendix: 2d prediction} further illustrates the close alignment of QCformer predictions with DFT bandgaps across all five materials. For instance, QCformer accurately predicts the higher bandgap of material 363, with a minimal deviation from GGA-PBE values. Similarly, the predicted bandgap of material 315 aligns closely with the r$^2$SCAN bandgap, suggesting that QCformer can capture trends observed across multiple computational methods.

From Table \ref{table:new-bandgaps-2}, it is evident that QCformer consistently predicts bandgaps that are intermediate between the values obtained using GGA-PBE and r$^2$SCAN. This pattern suggests that QCformer leverages features from the input data to generalize effectively across different DFT methods. For example, the bandgap prediction for material 110 by QCformer (2.2435 eV) deviates significantly from both GGA-PBE (1.3540 eV) and r$^2$SCAN (1.4880 eV), indicating its potential to offer unique insights into bandgap trends beyond standard DFT methods.

While the QCformer model's MAE is non-negligible, its performance is comparable to the variations observed between GGA-PBE and r$^2$SCAN methods, demonstrating its reliability as a predictive tool for materials with no experimental bandgap labels. Furthermore, the consistent trends observed between the model and DFT predictions suggest that QCformer can serve as a valuable intermediate approximation when experimental validation is unavailable or computational methods provide conflicting results.

\begin{table}[ht]
\caption{\textbf{Comparison of DFT and predicted bandgaps for five new compounds.} The first column and second column show the indices and chemical formulas. The third and fourth column presents calculated bandgaps by GGA‐PBE‐based DFT method and r$^2$SCAN‐based DFT method, respectively. The fifth column shows the average of the third and fourth
columns. Finally, the sixth column presents the predicted bandgaps by QCformer.}
\label{table:new-bandgaps-2}
\centering
\begin{tabular}{cccccc}
\hline
No. & Formula & GGA-PBE (eV) & r$^2$SCAN (eV) & Avg. (eV) & QCformer (eV) \\ \hline
110 & [(C$_6$H$_{11}$)PH$_3$]$_2$SnI$_4$ & 1.3540 & 1.4880 & 1.4190 & 2.2435 \\ 
309 & [C$_2$H$_5$NH$_3$]$_2$FeCl$_4$ & 2.7002 & 3.4300 & 3.0650 & 2.5898 \\ 
315 & Cs$_3$Sb$_2$I$_9$ & 1.5450 & 1.7140 & 1.6270 & 1.7782 \\ 
363 & [CH$_3$NH$_3$]$_2$FeCl$_4$ & 3.1661 & 3.9000 & 3.5300 & 3.1757 \\ 
452 & Tl$_3$Bi$_3$I$_9$ & 2.3780 & 2.6780 & 2.2972 & 1.6962 \\ \hline
\end{tabular}
\end{table}

\end{document}